\documentclass[sigconf]{acmart}
\renewcommand\footnotetextcopyrightpermission[1]{} 
\usepackage{xspace}
\usepackage{subfig}
\usepackage{tabu}
\usepackage{tabularx}
\usepackage{relsize}
\usepackage{float}
\usepackage{graphicx}
\usepackage{amsfonts}
\usepackage{amsmath}
\usepackage{enumitem}
\usepackage{adjustbox}
\usepackage{multirow}

\acmConference[KDD]{KDD}{2021}{Singapore}

\newif\ifshowcomments
\showcommentstrue

\ifshowcomments
\newcommand{\mynote}[2]{\fbox{\bfseries\sffamily\scriptsize{#1}}
{\small$\blacktriangleright$\textsf{\emph{#2}}$\blacktriangleleft$}}
\else
\newcommand{\mynote}[2]{}
\fi

\newcommand{\algo}{CTAB-GAN\xspace}

\newcommand{\ctgan}{CTGAN\xspace}
\newcommand{\tablegan}{TableGAN\xspace}
\newcommand{\cwgan}{CWGAN\xspace}
\newcommand{\medgan}{MedGAN\xspace}
\newcommand{\encoder}{Mixed-type Encoder\xspace}
\newcommand{\imbalanced}{imbalanced\xspace}
\newcommand{\skewed}{skewed\xspace}
\newcommand{\Skewed}{Skewed\xspace}
\newcommand{\column}{variable\xspace}
\newcommand{\columns}{variables\xspace}
\newcommand{\variable}{variable\xspace}
\newcommand{\variables}{variables\xspace}

\newcommand{\bimodal}{mixed\xspace}
\newcommand{\mixed}{mixed\xspace}
\newcommand{\Mixed}{Mixed\xspace}

\newcolumntype{L}[1]{>{\raggedright\let\newline\\\arraybackslash\hspace{0pt}}m{#1}}
\newcolumntype{C}[1]{>{\centering\let\newline\\\arraybackslash\hspace{0pt}}m{#1}}
\newcolumntype{R}[1]{>{\raggedleft\let\newline\\\arraybackslash\hspace{0pt}}m{#1}}

\newif\ifshowntationtable
\showntationtablefalse

\newif\ifnotdoubleblind
\notdoubleblindtrue

\begin{document}

\title{\algo: Effective Table Data Synthesizing}


\ifnotdoubleblind

\author{Zilong Zhao}
\authornotemark[1]
\email{Z.Zhao-8@tudelft.nl}
\author{Aditya Kunar}
\authornote{Both authors contributed equally to this research.}
\email{A.Kunar@student.tudelft.nl}

\affiliation{
  \institution{TU Delft}
  \city{Delft}
  \country{Netherlands}
}
  
\author{Hiek Van der Scheer}
\affiliation{%
  \institution{Aegon}
  \city{Den Haag}
  \country{Netherlands}}
\email{hiek.vanderscheer@aegon.com}  
  
\author{Robert Birke}
\affiliation{%
  \institution{ABB Research Switzerland}
  \city{D\"{a}ttwil}
  \country{Switzerland}}
\email{robert.birke@ch.abb.com}  

\author{Lydia Y. Chen}
\affiliation{%
  \institution{Tu Delft}
  \city{Delft}
  \country{Netherlands}}
\email{Y.Chen-10@tudelft.nl}

\fi


\begin{abstract}
While data sharing is crucial for knowledge development,  privacy concerns and strict regulation (e.g., European General Data Protection Regulation (GDPR)) unfortunately limit its full effectiveness. Synthetic tabular data emerges as an alternative to enable data sharing while fulfilling regulatory and privacy constraints. 
The state-of-the-art tabular data synthesizers draw methodologies from Generative Adversarial Networks (GAN) and address two main data types in industry, i.e., continuous and categorical. In this paper, we develop \algo, a novel conditional table GAN architecture that can effectively model diverse data types, including a mix of continuous and categorical \variables. 
Moreover, we address data imbalance and long tail issues, i.e., certain \variables have drastic frequency differences across large values. 
To achieve those aims, we first introduce the information loss and classification loss to the conditional GAN. Secondly, we design a novel conditional vector, which efficiently encodes the mixed data type and skewed distribution of data \variable. We extensively evaluate \algo with the state of the art GANs that generate synthetic tables, in terms of data similarity and analysis utility. The results on five datasets show that the synthetic data of \algo remarkably resembles the real data for all three types of \variables and results into higher accuracy for five machine learning algorithms, by up to 17\%.
\end{abstract}




\keywords{GAN, data synthesis, tabular data, imbalanced distribution}


\maketitle

\section{Introduction}

``Data is the new oil'' is a quote that goes back to 2006, which is credited to mathematician Clive Humby. It has recently picked up more steam after The Economist published a 2017 report~\cite{theeconomist} titled ``The world's most valuable resource is no longer oil, but data''. Many companies nowadays discover valuable business insights from various internal and external data sources. However, the big knowledge behind big data often impedes personal privacy and leads to unjustified analysis~\cite{narayanan2008}.  To prevent the abuse of data and the risks of privacy breaching, the European Commission introduced the European General Data Protection Regulation (GDPR)  and enforced strict data protection measures. This however instills a new challenge in the data-driven industries to look for new scientific solutions that can empower big discovery while respecting the constraints of data privacy and governmental regulation.  

An emerging solution is to leverage synthetic data~\cite{cramergan}, which statistically resembles real data and can comply with GDPR due to its synthetic nature. The industrial datasets (at stakeholders like banks,  insurance companies, and health care) present multi-fold challenges. First of all, such datasets are organized in tables and populated with both continuous and categorical variables, or a mix of the two, e.g., the value of mortgage for a loan holder. This value can be either 0 (no mortgage) or some continuous positive number. Here, we term such a type of \variables, \textbf{\mixed \variable}.  Secondly data \variables often have a wide range of values as well as skewed frequency distribution, e.g., the statistic of transaction amount for credit card. Most transactions should be within 0 and 500 bucks (i.e. daily shopping for food and clothes), but exceptions of a high transaction amount surely exist.

Generative Adversarial Network (GAN)~\cite{gan} is one of the emerging data synthesizing methodologies. The GAN is first trained on a real dataset. Then used to generate data. 
Beyond its success on generating images, GAN has recently been applied to generate tabular data~\cite{tgan, ctgan, tablegan, cramergan}. The state of the art of tabular generators~\cite{ctgan} treats categorical variables via conditional GAN, where each categorical value is considered as a condition. However, their focus is only on two types of variables, namely continuous and categorical, overlooking an important class of \mixed data type. In addition, it is unclear if existing solutions can efficiently handle highly \imbalanced categorical variables and skewed continuous variables.


In this paper, we aim to design a tabular data synthesizer that addresses the limitations of the prior state-of-the-art: (i) encoding \mixed data type of continuous and categorical variables, (ii) efficient modeling of long tail continuous variables and (iii) increased robustness to \imbalanced categorical \variables along with skewed continuous \variables. Hence, we propose a novel conditional table generative adversarial network, \algo. Two key features of \algo are the introduction of classification loss in conditional GAN, and novel encoding for the conditional vector that efficiently encodes mixed variables and helps to deal with highly skewed distributions for continuous variables.




\begin{figure*}[t]
	\begin{center}
		\subfloat[Mortgage in Loan dataset~\cite{kaggleloan}]{
			\includegraphics[width=0.33\textwidth]{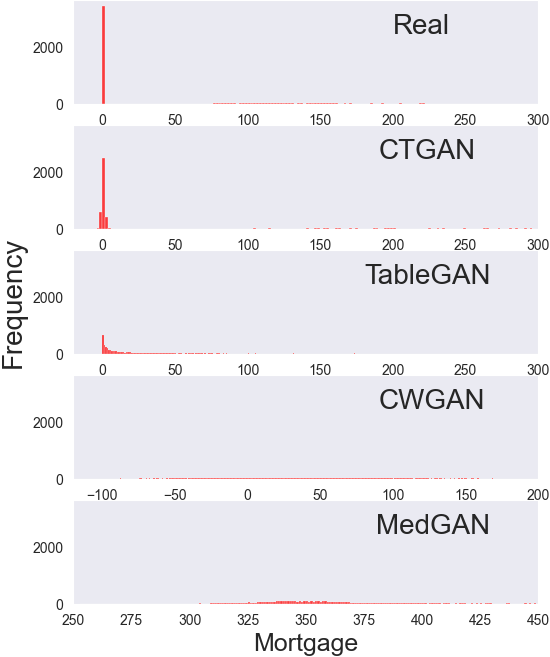}
			\label{fig:mortgage_column_motivation}
		}
		\subfloat[Amount in Credit dataset~\cite{UCIdataset}]{
			\includegraphics[width=0.3\textwidth]{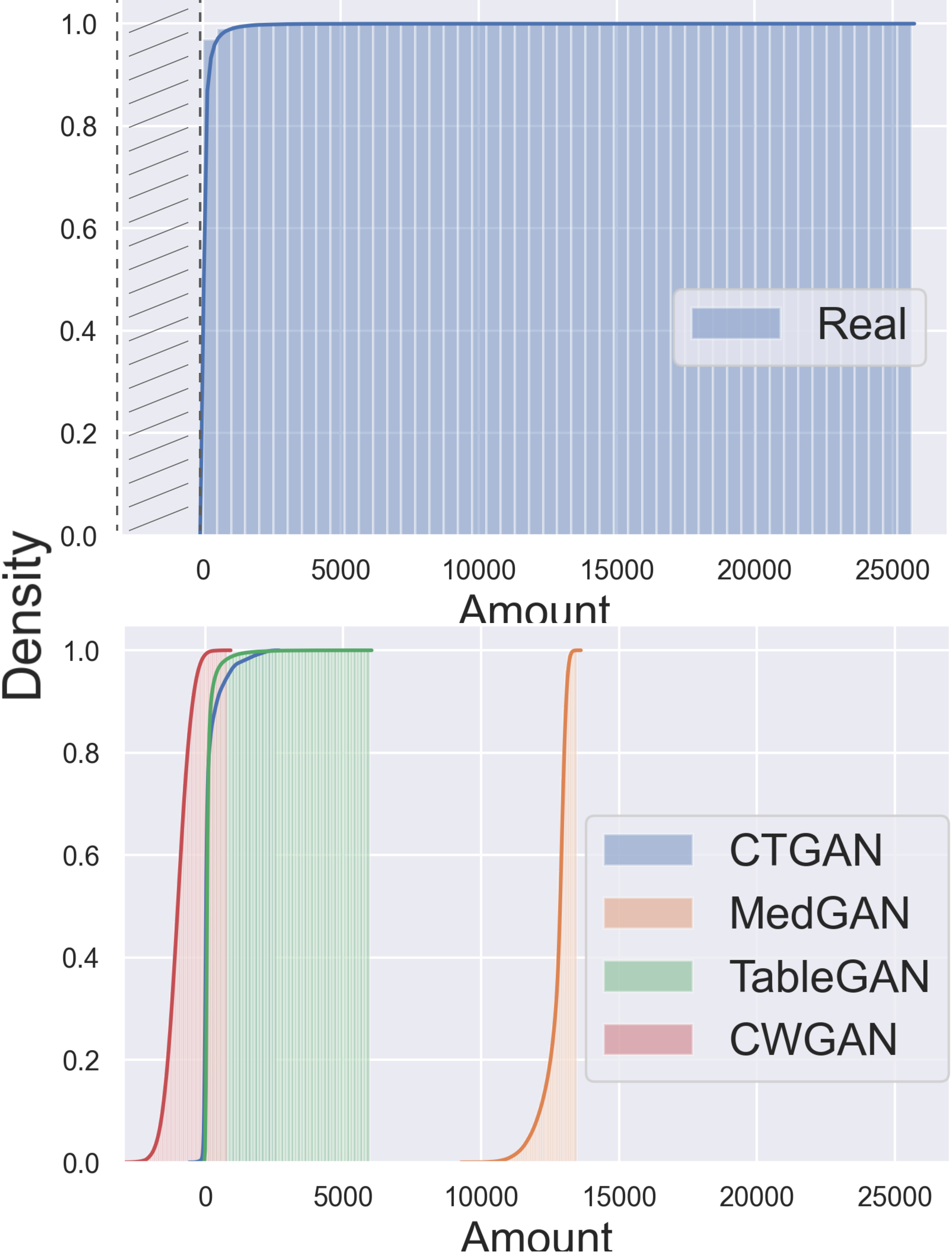}
			\label{fig:amount_result_motivation}
		}
		\subfloat[Hours-per-week in Adult dataset~\cite{UCIdataset}]{
			\includegraphics[width=0.335\textwidth]{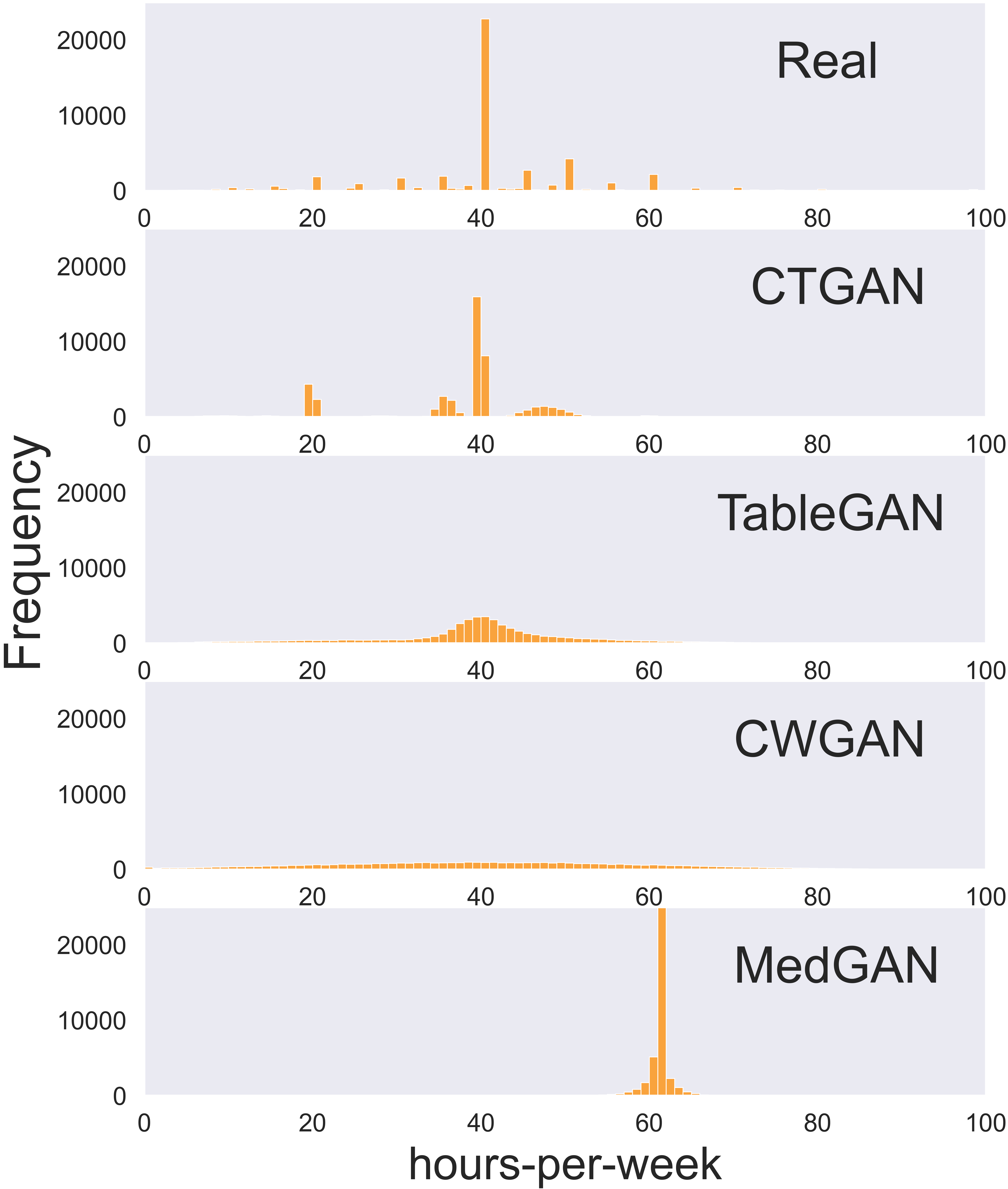}
			\label{fig:gmm_result_motivation}
		}
		\caption{ Challenges of modeling industrial dataset using existing GAN-based table generator: (a) \bimodal type, (b) long tail distribution, and (c) \skewed data} 
		\label{fig:motivationcases}
 	\end{center}
 \vspace{-1em}
\end{figure*}

We rigorously evaluate \algo in three dimensions: (i) utility of machine learning based analysis on the synthetic data, (ii) statistical similarity to the real data, and (iii) privacy preservability.
Specifically, the proposed \algo is tested on 5 widely used machine learning datasets: Adult, Covertype, Credit, Intrusion and Loan, against 4 state-of-the-art GAN-based tabular data generation algorithms: \ctgan, \tablegan, \cwgan and \medgan. Our results show that \algo not only outperforms all the comparisons in machine learning utility and statistical similarity but also provides better distance-based privacy guarantees than \tablegan, the second best performing algorithm in the machine learning utility and statistical similarly evaluation.
 
The main contributions of this study can be summarized as follows:
\begin{itemize}[noitemsep]

 \item Novel conditional adversarial network which introduces a classifier providing additional supervision to improve its utility for ML applications. 
 
 \item Efficient modelling of continuous, categorical, and \textbf{\bimodal} variables via novel data encoding and conditional vector.
 
 \item Light-weight data pre-processing to mitigate the impact of long tail distribution of continuous variables. 
 
 \item Providing an effective data synthesizer for the relevant stakeholders.

 
\end{itemize}

\subsection{Motivation}
\label{sec:motivation}

In this subsection, we empirically demonstrate how the prior state-of-the-art methods fall short in solving challenges in industrial data sets. The detailed experimental setup can be found in Sec.~\ref{ssec:setup}.

\textbf{\Mixed data type \columns}. To the best of our knowledge, existing GAN-based tabular generators only consider table columns as either categorical or continuous. However, in reality, a \column can be a mix of these two types, and often \columns have missing values. 
The \textit{Mortgage} \column from the Loan dataset is a good example of \bimodal variable.
Fig.~\ref{fig:mortgage_column_motivation} shows the distribution of the original and synthetic data generated by 4 state-of-the-art algorithms for this \column.
According to the data description, a loan holder can either have no mortgage (0 value) or a mortgage (any positive value). In appearance this \column is not a categorical type due to the numeric nature of the data. So all 4 state-of-the-art algorithms treat this \columns as continuous type without capturing the special meaning of the value zero. Hence, all 4 algorithms generate a value around 0 instead of exact 0. And the negative values for Mortgage have no/wrong meaning in the real world.

\textbf{Long tail distributions}. Many real world data can have long tail distributions where most of the occurrences happen near the initial value of the distribution, and rare cases towards the end.
Fig.~\ref{fig:gmm_result_motivation} plots the cumulative frequency for the original (top) and synthetic (bottom) data generated by 4 state-of-the-art algorithms for the
\textit{Amount} in the Credit dataset. This \column represents the transaction amount when using credit cards. One can imagine that most transactions have small amounts, ranging from few bucks to thousands of dollars. However, there definitely exists a very small number of transactions with large amounts. Note that for ease of comparison both plots use the same x-axis, but Real has no negative values. 
Real data clearly has 99\% of occurrences happening at the start of the range, but the distribution extends until around $25000$. In comparison none of the synthetic data generators is able to learn and imitate this behavior.
    

\textbf{\Skewed multi-mode continuous variables}.
The term \textit{multi-mode} is extended from Variational Gaussian Mixtures (VGM). More details are given in Sec.~\ref{ssec:data_representation}. The intuition behind using multiple modes can be easily captured from Fig.~\ref{fig:gmm_result_motivation}. The figure plots in each row the distribution of the working \textit{Hours-per-week} \column from the Adult dataset.
This is not a typical Gaussian distribution.
There is an obvious peak at 40 hours but with several other lower peaks, e.g. at 50, 20 and 45. 
Also the number of people working 20 hours per week is higher than those working 10 or 30 hours per week.
This behavior is difficult to capture for the state-of-the-art data generators (see subsequent rows in Fig.\ref{fig:gmm_result_motivation}). The closest results are obtained by \ctgan which uses Gaussian mixture estimation for continuous variables. However, \ctgan loses some modes compared to the original distribution.
    

The above examples show the shortcomings of current state-of-the-art GAN-based tabular data generation algorithms 
and motivate the design of our proposed \algo.

\section{Related Studies}
We divide the related studies using GAN to generate tabular data into two categories: (i) based on GAN, and (ii) based on conditional GAN.
\textbf{GAN-based generator} Several studies extend GAN to accommodate categorical variables by augmenting GAN architecture. 
MedGAN~\cite{medgan} combines an auto-encoder with a GAN. It can generate continuous or discrete variables, and has been applied to generate synthetic electronic health record (EHR) data. CrGAN-Cnet~\cite{cramergan} uses GAN to conduct Airline Passenger Name Record Generation. It integrates the Cramér Distance~\cite{cramerdistance} and Cross-Net architecture~\cite{crossnet} into the algorithm. In addition to generating with continuous and categorical data types,  CrGAN-Cnet can also handle missing value in the table by adding new \columns. TableGAN~\cite{tablegan} introduces information loss and a classifier into GAN framework. It specifically adopts Convolutional Neural Network (CNN) for generator, discriminator and classifier. 
Although aforementioned algorithms can generate tabular data, they cannot specify how to generate from a specific class for particular variable. For example, it is not possible to generate health record for users whose sex is female. 
In addition to data generation, privacy is another important factor for synthetic tabular data. PATE-GAN~\cite{pategan} is not specifically designed for tabular data generation, but it proposes a framework which generates  synthetic  data  with  differential  privacy  guarantees.

\textbf{Conditional GAN-based generator} Due to the limitation of controlling generated data via GAN, Conditional GAN is increasingly used,  and its conditional vector can be used to specify to generate a particular class of data.  CW-GAN~\cite{cwgan} applies the Wasserstein distance~\cite{wgan} into the conditional GAN framework. It leverages the usage of conditional vector to oversample the minority class to address imbalanced tabular data generation. CTGAN~\cite{ctgan} integrates PacGAN~\cite{pacgan} structure in its discriminator and uses WGAN loss plus gradient penalty~\cite{wgan_gp} to train a conditional GAN framework. It also adopts a strategy called training-by-sampling, which takes advantage of conditional vector, to deal with the imbalanced categorical variable problem. 

In our paper, we not only focus on modelling continuous or categorical variables, but also cover the \mixed data type (i.e., \columns that contain both categorical and continuous values, or even missing values). We effectively combine the strengths of prior art, such as classifier, information loss, effective encoding, and conditional vector. Furthermore, we proactively address the pain point of long tail \variable distributions 
and propose a new conditional vector structure to better deal with imbalanced datasets.

\section{\algo}

\begin{figure*}[t]
    \centering
    \includegraphics[width=0.95\linewidth]{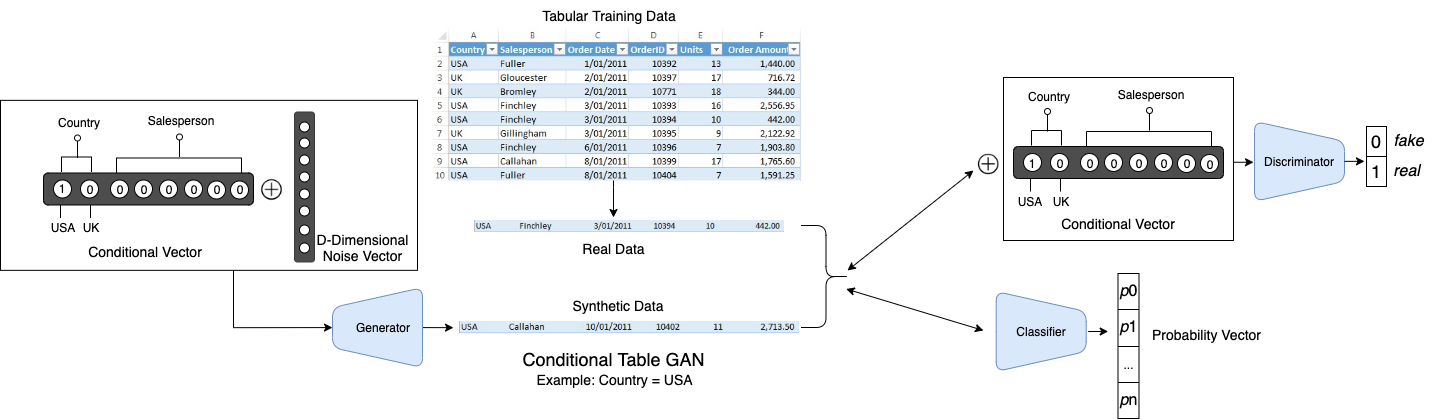}
    \caption{Synthetic Tabular Data Generation via \algo}
    \label{fig:STD}
   \vspace{-0.5em}
\end{figure*}

\algo is a tabular data generator designed to overcome the challenges outlined in Sec.~\ref{sec:motivation}. In \algo we invent a \textit{\encoder} which can better represent mixed categorical-continuous variables as well as missing values. \algo is based on a conditional GAN (CGAN) to efficiently treat minority classes~\cite{cwgan,ctgan}, with the addition of classification and information loss~\cite{tablegan,acgan} to improve semantic integrity and training stability, respectively. Finally, we leverage a log-frequency sampler to overcome the mode collapse problem for imbalanced \columns.   
\ifshowntationtable
Before illustrating \algo, we summarized the used notation in Table~\ref{tab:symboltable}.

\begin{table}
\centering
\caption{Notation}
\label{tab:symboltable}
\begin{tabular}{|C{1.5cm} |L{6cm}|} \hline
\textbf{Symbol}	& \textbf{Description}\\ \hline
$\mathcal{G}$	& Generator of \algo					 \\\hline
$\mathcal{D}$ 	& Discriminator of \algo					 \\\hline
$\mathcal{C}$	& Classifier of \algo					 \\\hline
$\mathcal{V}$	& Conditional vector 
\\\hline
$\mathcal{N}$ 	& Normal distribution					 \\\hline
$\mu_i$ 	& value of categorical mode, mean value of Gaussian mixture mode					 \\\hline
$\rho_i$ 	& probability density calculated based on mode $i$					 \\\hline
$\gamma_{i,j}$&  one-hot encoding vector for $i$th row value in categorical \column $j$			\\\hline
$\bigoplus$&  vector concatenation			\\\hline
\end{tabular}
\end{table}
\fi

\subsection{Technical background}

GANs are a popular method to generate synthetic data first applied with great success to images~\cite{stylegan,stylegan2} and later adapted to tabular data~\cite{yahi_gan}. GANs leverage an adversarial game between a generator trying to synthesize realistic data and a discriminator trying to discern synthetic from real samples.

To address the problem of dataset imbalance, we leverage \textit{conditional generator} and \textit{training-by-sampling} methods~\cite{ctgan}. The idea behind this is to use an additional vector, termed as the conditional vector, to represent the classes of categorical \columns. This vector is both fed to the generator and used to bound the sampling of the real training data to subsets satisfying the condition. We can leverage the condition to resample all classes giving higher chances to minority classes to train the model. 

To enhance the generation quality, we incorporate two extra terms in the loss function of the generator~\cite{tablegan,acgan}: information and classification loss. 
The information loss penalizes the discrepancy between statistics of the generated data and the real data. This helps to generate data which is statistically closer to the real one.
The classification loss requires to add to the GAN architecture an auxiliary classifier in parallel to the discriminator. 
For each synthesized label the classifier outputs a predicted label.
The classification loss quantifies the discrepancy between the synthesized and predicted class. This helps to increase the semantic integrity of synthetic records.
For instance, (sex=female, disease=prostate cancer) is not a semantically correct record as women do not have a prostate, and no such record should appear in the original data and is hence not learnt by the classifier.

To counter complex distributions in continuous \columns we embrace the \textit{mode-specific normalization} idea~\cite{ctgan} which encodes each value as a value-mode pair stemming from Gaussian mixture model.



\subsection{Design of \algo}
\label{sec:design_algorithm}
The structure of \algo comprises three blocks: Generator $\mathcal{G}$, Discriminator $\mathcal{D}$ and an auxiliary Classifier $\mathcal{C}$ (see Fig.~\ref{fig:STD}). 
Since our algorithm is based on conditional GAN, the generator requires a noise vector plus a conditional vector. Details on the conditional vector are given in Sec.~\ref{sec:condvec}. To simplify the figure, we omit the encoding and decoding of the synthetic and real data detailed in Sec.~\ref{ssec:data_representation}.

GANs are trained via a zero-sum minimax game where the discriminator tries to maximize the objective, while the generator tries to minimize it. The game can be seen as a mentor ($\mathcal{D}$) providing feedback to a student ($\mathcal{G}$) on the quality of his work.
Here, we introduce additional feedback for $\mathcal{G}$ based on the information loss and classification loss. The information loss matches the first-order (i.e., mean) and second-order (i.e., standard deviation) statistics of synthesized and real records. This leads the synthetic records to have the same statistical characteristics as the real records.
In addition, the classifier is trained to learn the correlation between classes and the other \column values using the real training data.
The classification loss helps to check the semantic integrity, and penalizes synthesized records where the combination of values are semantically incorrect.
These two losses are added to the original loss term of $\mathcal{G}$ during training.

$\mathcal{G}$ and $\mathcal{D}$ are implemented by a four and a two layers CNN, respectively.
CNNs are good at capturing the relation between pixels within an image~\cite{lecuncnn}, which in our case, can help to increase the semantic integrity of synthetic data.
$\mathcal{C}$ uses a 7 layers MLP. The classifier is trained on the original data to better interpret the semantic integrity. Hence synthetic data are reverse transformed from their encoding (details in Sec.~\ref{ssec:data_representation}) before being used as input for $\mathcal{C}$ to create the class label predictions.

\subsection{\encoder}
\label{ssec:data_representation}

\begin{figure}[t]
	\begin{center}
		\subfloat[Mixed type \column distribution with VGM]{
			\includegraphics[width=0.47\columnwidth]{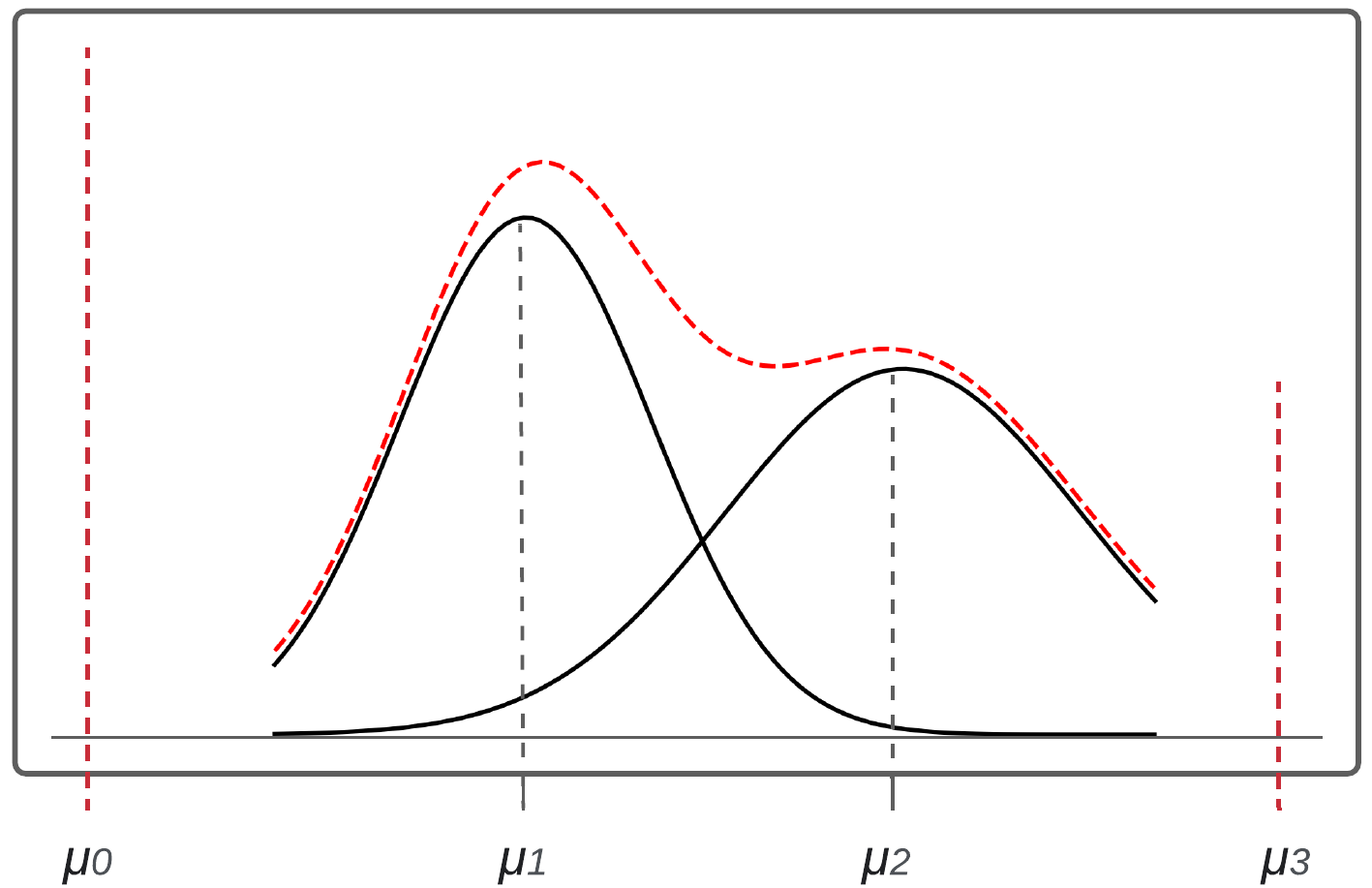}
			\label{fig:vgm}
		}
		\hfil
		\subfloat[Mode selection of single value in continuous \column]{
			\includegraphics[width=0.47\columnwidth]{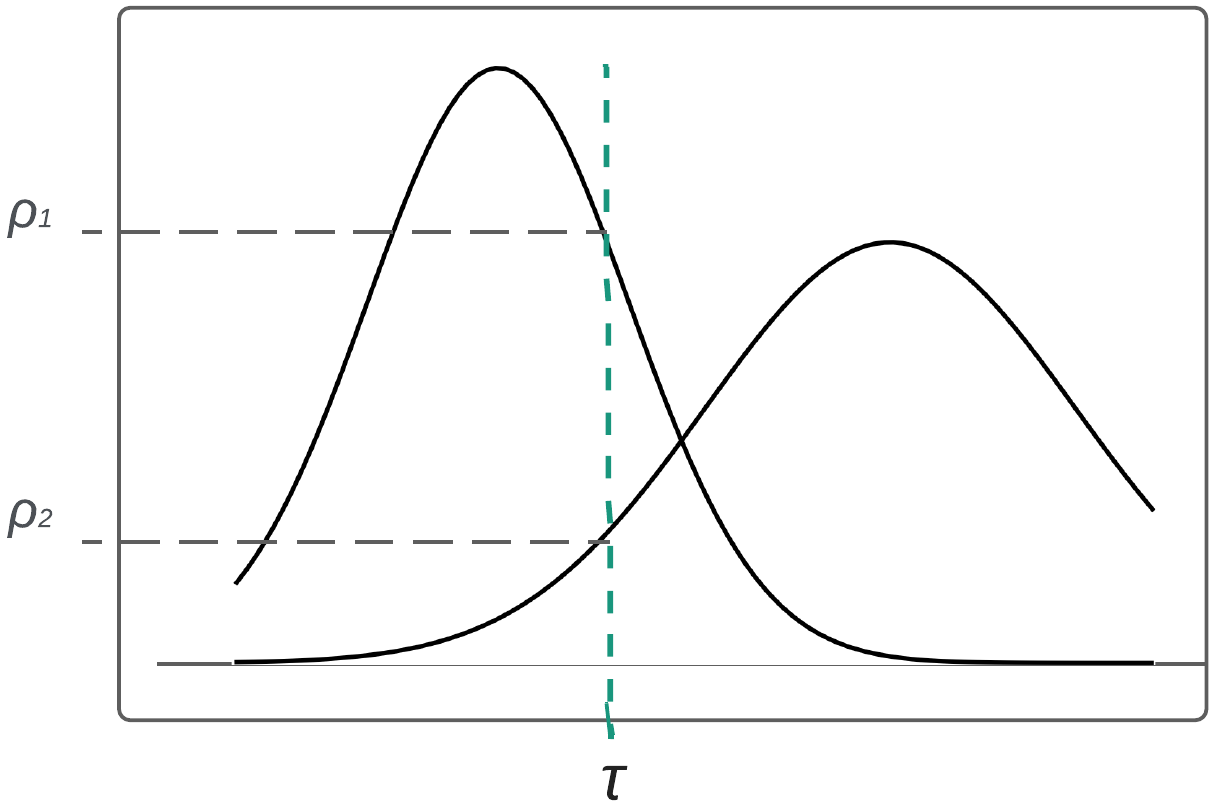}
			\label{fig:vgm_single}
		}
		\caption{Encoding for mix data type \column}
		\label{fig:gmm_distribution}
	\end{center}
	\vspace{-0.5em}
\end{figure}

The tabular data is encoded \column by \column. We distinguish three types of \columns: categorical, continuous and \mixed. 
We define \columns as mixed if they contain both categorical and continuous values or continuous values with missing values. We propose the new \encoder to deal with such \columns. With this encoder, values of \mixed \columns are seen as concatenated value-mode pairs. We illustrate the encoding via the exemplary distribution of a \mixed \column shown in red in Fig.~\ref{fig:vgm}. One can see that values can either be exactly $\mu_0$ or $\mu_3$ (the categorical part) or distributed around two peaks in $\mu_1$ and $\mu_2$ (the continuous part). We treat the continuous part using a variational Gaussian mixture model (VGM)~\cite{prml} to estimate the number of modes $k$, e.g. $k=2$ in our example, and fit a Gaussian mixture. The learned Gaussian mixture is:
\begin{equation}
    \mathbb{P} = \sum_{k=1}^{2} \omega_k \mathcal{N}(\mu_k, \sigma_k)
\end{equation}
where $\mathcal{N}$ is the normal distribution and $\omega_k$, $\mu_k$ and $\sigma_k$ are the weight, mean and standard deviation of each mode, respectively.

To encode values in the continuous region of the \column distribution, we associate and normalize each value with the mode having the highest probability (see Fig.~\ref{fig:vgm_single}). Given $\rho_1$ and $\rho_2$ being the probability density from the two modes in correspondence of the \column value $\tau$ to encode, we select the mode with the highest probability. In our example $\rho_1$ is higher and we use mode $1$ to normalize $\tau$. The normalized value $\alpha$ is:
\begin{equation}
    \alpha =  \frac{\tau - \mu_1}{4\sigma_1}
\end{equation}
Moreover we keep track of the mode $\beta$ used to encode $\tau$ via one hot encoding, e.g.  $\beta = [0,1,0,0]$ in our example. The final encoding is giving by the concatenation of $\alpha$ and $\beta$: $\alpha \bigoplus \beta$
\ifshowntationtable
.
\else
where $\bigoplus$ is the vector concatenation operator.
\fi

The categorical values are treated similarly, except $\alpha$ directly represents the value of the mode, e.g. corresponding to $\mu_0$ or $\mu_3$ in our example. Hence, for a value in $\mu_3$, the final encoding is given by $\mu_3 \bigoplus [0, 0, 0, 1]$. Note that categorical values are not limited to numbers. They can be of any type such as a string or even missing. We can map these symbols to a numeric value outside of the range of the continuous region.

Categorical \columns use the same encoding as the continuous intervals of \mixed \columns. Categorical \columns are encoded via a one-hot vector $\gamma$. Missing values are treated as a separate unique class and we add an extra bit to the one-hot vector for it. A row with $[1, \dots, N]$ \columns is encoded by concatenation of the encoding of all \column values, i.e. either $(\alpha \bigoplus \beta)$ for continuous and mixed \columns or $\gamma$ for categorical \columns. 
Having $n$ continuous/mixed \columns and $m$ categorical \columns ($n + m = N$) the final encoding is: 
\begin{equation}
\label{condvec}
    \bigoplus_{i=1}^{n} \alpha_i\mathsmaller{\bigoplus} \beta_{i} \;
    \bigoplus_{j=n+1}^{N} \gamma_{j} 
\end{equation}

\subsection{Counter \imbalanced training datasets}
\label{sec:condvec}

In \algo, we use conditional GAN to counter imbalanced training datasets.
When we sample real data, we use the conditional vector to filter and rebalance the training data.

\begin{figure}[t]
    \centering
    \includegraphics[scale=.18]{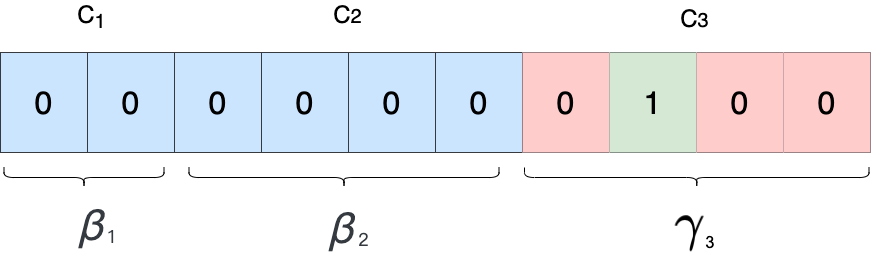}
    \caption{Conditional vector: example selects class 2 from third \column out of three}
    \label{fig:condvec}
     \vspace{-1em}
\end{figure}

The conditional vector $\mathcal{V}$ is a bit vector given by the concatenation of all mode one-hot encodings $\beta$ (for continuous and \mixed \columns) and all class one-hot encodings $\gamma$ (for categorical \columns) for all \columns present in Eq.~\eqref{condvec}. Each conditional vector specifies a single mode or a class. More in detail, $\mathcal{V}$ is a zero vector with a single one in correspondence to the selected \column with selected mode/class.
Fig.~\ref{fig:condvec} shows an example with three \columns, one continuous ($C_1$), one \mixed ($C_2$) and one categorical ($C_3$), with class 2 selected on $C_3$. 

To rebalance the dataset, each time we need a conditional vector during training, we first randomly choose a \column with uniform probability. Then we calculate the probability distribution of each mode (or class for categorical \columns) in that \column using frequency as proxy and sample a mode based on the logarithm of its probability. Using the log probability instead of the original frequency gives minority modes/classes higher chances to appear during training. This helps to alleviate the collapse issue for rare modes/classes.

\subsection{Treat long tail}

We encode continuous values using variational Gaussian mixtures to treat multi-mode data distributions (details in Sec.~\ref{ssec:data_representation}).
However, Gaussian mixtures can not deal with all types of data distribution, notably distributions with long tail where few rare points are far from the bulk of the data.
VGM has difficulty to encode the values towards the tail.
To counter this issue we pre-process \columns with long tail distributions with a logarithm transformation. For such a \column having values with lower bound $l$, we replace each value $\tau$ with compressed $\tau^c$:
\begin{equation}
\label{eq:preprossesing}
\tau^c =  \left\{
	\begin{array}{rl}
		 \mbox{log($\tau$)} &  \mbox{if $l>$0} \\
		\mbox{log($\tau$ - $l$+$\epsilon$)} & \mbox{if $l\leqslant$0}  \mbox{, where } \mbox{$\epsilon>$0} 	 
	\end{array} \right\}
\end{equation}

The log-transform allows to compress and reduce the distance between the tail and bulk data making it easier for VGM to encode all values, including tail ones. We show the effectiveness of this simple yet performant method in Sec.~\ref{sec:further}.
\section{Experimental Analysis}

To show the efficacy of the proposed \algo, we select five commonly used machine learning datasets, and compare with four state-of-the-art GAN based tabular data generators. We evaluate the effectiveness of \algo in terms of the resulting ML utility, statistical similarity to the real data, and privacy distance.  Moreover, we provide an ablation analysis to highlight the efficacy of the unique components of \algo.

\subsection{Experimental setup}
\label{ssec:setup}

 {\bf Datasets}. Our algorithm is tested on five commonly used machine learning datasets. Three of them -- \textbf{Adult}
 , \textbf{Covertype}
 and \textbf{Intrusion}
 -- are from the UCI machine learning repository~\cite{UCIdataset}. The other two --  \textbf{Credit}\cite{kagglecredit}
 and \textbf{Loan}\cite{kaggleloan}
 -- are from Kaggle. All five tabular datasets have a target \column, for which we use the rest of the \columns to perform classification. Due to computing resource limitations, 50K rows of data are sampled randomly in a stratified manner with respect to the target \column for Covertype, Credit and Intrusion datasets.  
 However, the Adult and Loan datasets are not sampled. The details of each dataset are shown in Tab.~\ref{table:DD}. 
\begin{table}[t]
\centering
\caption{Description of datasets. The notations $C$, $B$, $M$ \& $Mi$ represent number of continuous, binary, multi-class categorical, \mixed \variables and imbalance ratios (i.e no. of minority samples/no. of majority samples) respectively.}
\begin{tabular}{ |C{1.2cm}|C{1.4cm}|C{1.6cm}|C{0.4cm}|C{0.4cm}|C{0.4cm}|C{0.6cm}|C{0.6cm}| }
\hline
\textbf{Dataset} & \textbf{Train/Test Split} &\textbf{Target \column} & \textbf{\small$\mbox{\#C}$}  & \textbf{\small$\mbox{\#B}$} & \textbf{\small$\mbox{\#M}$} & \textbf{\small$\mbox{\#Mi}$}\\ 
\hline
\small{Adult}     & 39k/9k   &\small "income"        & 3   & 2  & 7  & 2\\
\hline
 \small Covertype & 45k/5k    &\small "Cover\_Type"    & 9   & 44 & 1  & 1 \\\hline
\small Credit    & 40k/10k    &\small "Class"         & 30  & 1  & 0  & 0\\ 
\hline
 \small Intrusion & 45k/5k    &\small "Class"         & 4   & 6  & 14 & 18 \\\hline
\small Loan      & 4k/1k       & "PersonalLoan"  & 5   & 5  & 2  & 1 \\
\hline
\end{tabular}
\label{table:DD}
\vspace{-0.8em}
\end{table}

{\bf Baselines}. Our \algo is compared with 4 state-of-the-art GAN-based tabular data generators: \ctgan, \tablegan, \cwgan and \medgan. To have a fair comparison, all algorithms are coded using Pytorch, with the generator and discriminator structures matching the descriptions provided in their respective papers. For Gaussian mixture estimation of continuous variables, we use the same settings as the evaluation of \ctgan, i.e. 10  modes. All algorithms are trained for 150 epochs for Adult, Covertype, Credit and Intrusion datasets, whereas the algorithms are trained for 300 epochs on Loan dataset. This is because, the Loan dataset is significantly smaller than the others containing only 5000 rows and requires a long training time to converge. 
Lastly, each experiment is repeated 3 times. 

{\bf Environment}. Experiments are run under Ubuntu 20.04 on a machine equipped with 32 GB memory, a GeForce RTX 2080 Ti GPU and a 10-core Intel i9 CPU.

\subsection{Evaluation metrics}
\label{sec:metrics}
The evaluation is conducted on three dimensions: (1) machine learning (ML) utility, (2) statistical similarity and (3) privacy preservability. The first two are used to evaluate if the synthetic data can be used as a good proxy of the original data. The third criterion sheds light on the nearest neighbour distances within and between the original and synthetic datasets, respectively. 
  
\subsubsection{Machine learning (ML) utility}
\label{sec:ml_efficacy}
\begin{figure}[t]
    \centering
    \includegraphics[width=0.99\columnwidth]{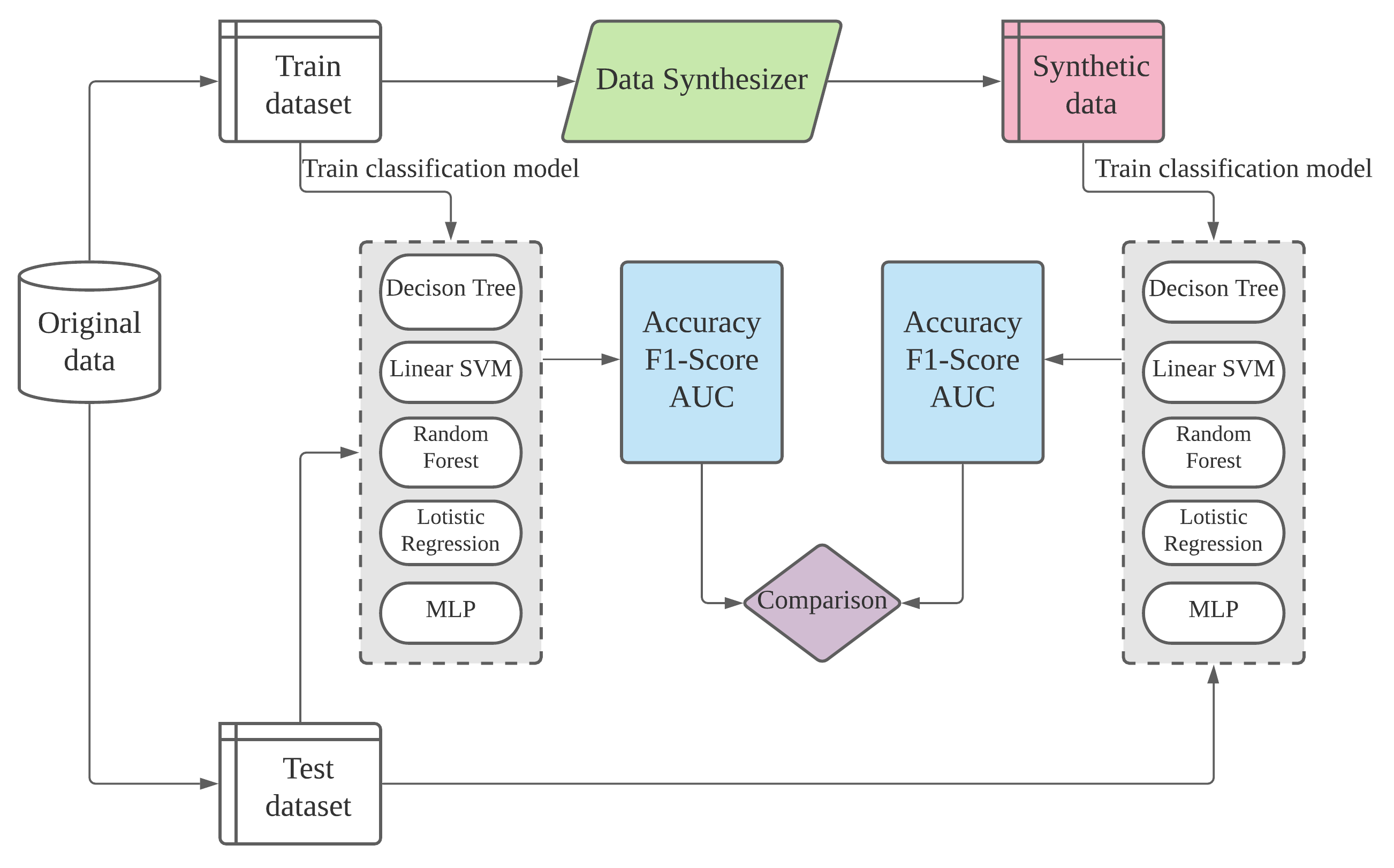}
    \caption{Evaluation flows for ML utility}
    \label{fig:settingA}
    \vspace{-0.5em}
\end{figure}

As shown in Fig.~\ref{fig:settingA}, to evaluate the ML utility of synthetic data, the original and synthetic data are evaluated by 5 widely used machine learning algorithms: decision tree classifier, linear support-vector-machine (SVM),  random forest classifier, multinomial logistic regression and  multi-layer-perceptron (MLP). 

We first split the original dataset into training and test datasets. The training set is  used as input to the GAN models as the real dataset. Once the training is finished, we use the GAN models to generate synthetic data with the same size as the training set. The synthetic and real training datasets are then separately used to train the above-mentioned machine learning models and evaluated on the real test datasets. The machine learning performance is measured via the accuracy, F1-score and area under the ROC. The aim of this design is to test how close the ML utility is when we train a machine learning model using the synthetic data vs the real data. This responds to the question, ``Can synthetic data be used as a proxy of the original data for training ML models?''. 

\subsubsection{Statistical Similarity}
Three metrics are used to quantitatively measure the statistical similarity between the real and synthetic data. 

\textbf{Jensen-Shannon divergence (JSD)}~\cite{jsd}. The JSD provides a measure to quantify the difference between the probability mass distributions of individual categorical \columns belonging to the real and synthetic datasets, respectively. Moreover, this metric is bounded between 0 and 1 and is symmetric allowing for an easy interpretation of results. 

\textbf{Wasserstein distance (WD)}~\cite{wgan_test}. In similar vein, the Wasserstein distance is used to capture how well the distributions of individual continuous/mixed \columns are emulated by synthetically produced datasets in correspondence to real datasets. 
We use WD because we found that the JSD metric was numerically unstable for evaluating the quality of continuous \columns, especially when there is no overlap between the synthetic and original dataset. Hence, we resorted to utilize the more stable Wasserstein distance. 
 
\textbf{Difference in pair-wise correlation (Diff. Corr.)}. 
To evaluate how well feature interactions are preserved in the synthetic datasets, we first compute the pair-wise correlation matrix for the columns within real and synthetic datasets individually. To measure the correlation between any two continuous features, the Pearson correlation coefficient is used. It ranges between $[-1,+1]$. Similarly, the Theil uncertainty coefficient is used to measure the correlation between any two categorical features. It ranges between $[0,1]$. Lastly, the correlation ratio  between categorical and continuous \columns is used. It also ranges between $[0,1]$. Note that the dython\footnote{\url{http://shakedzy.xyz/dython/modules/nominal/\#compute\_associations}} library is used to compute these metrics. Finally, the differences between the pair-wise correlation matrices for the real and synthetic datasets is computed.

\subsubsection{Privacy preservability}
To quantify the privacy preservability, we resort to distance metrics (instead of differential privacy~\cite{pategan}) as they are intuitive and easy to understand by data science practitioners. Specifically, the following two metrics are used to evaluate the privacy risk associated with synthetic datasets.

\textbf{Distance to Closest Record (DCR)}. The DCR is used to measure the Euclidean distance between any synthetic record and its closest corresponding real neighbour. Ideally, the higher the DCR the lesser the risk of privacy breach. Furthermore, the $5^{th}$ percentile of this metric is computed to provide a robust estimate of the privacy risk. 
    
\begin{figure}[t]
    \centering
    \includegraphics[width=0.8\columnwidth]{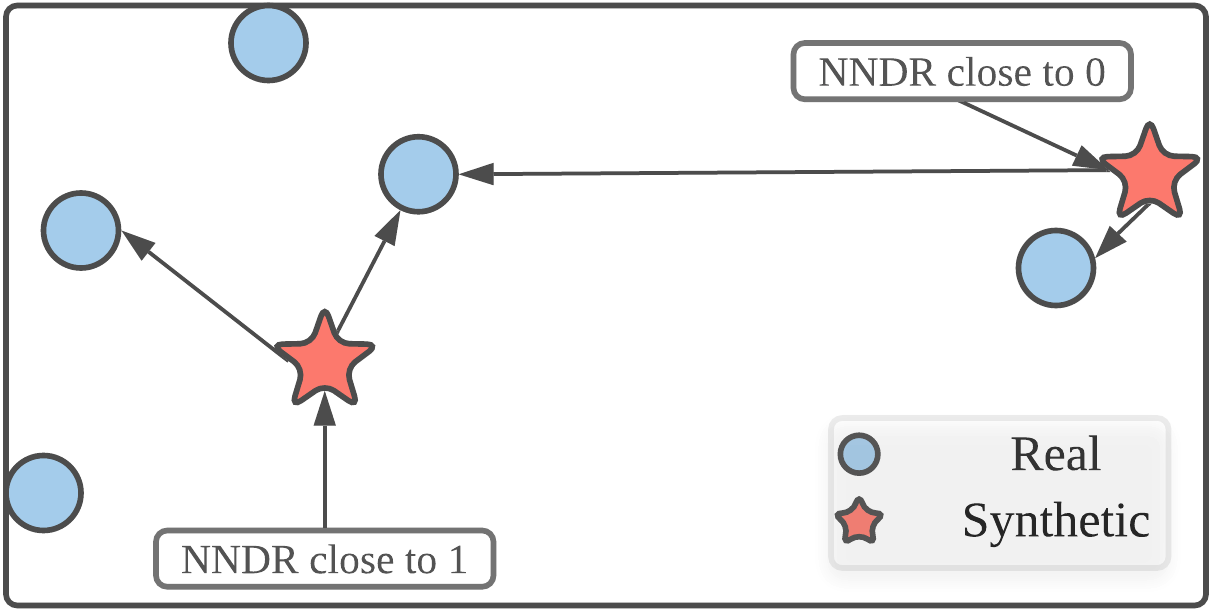}
    \caption{Illustration of NNDR metric with its privacy risk implications}
    \label{fig:nndr}
\end{figure}

\textbf{Nearest Neighbour Distance Ratio (NNDR)}~\cite{nndr}. Instead of only measuring the closest neighbour, the NNDR measures the ratio between the Euclidean distance for the closest and second closest real neighbour to any corresponding synthetic record. This ratio is within $[0,1]$. Higher values indicate better privacy. 
Low NNDR values between synthetic and real data may reveal sensitive information from the closest real data record. Fig.~\ref{fig:nndr} illustrates the case.  Hence, this ratio helps to evaluate the privacy risk with greater depth and better certainty. Note that the $5^{th}$ percentile is computed here as well.

\subsection{Results analysis}

\begin{figure*}[t]
 	\begin{center}
	\hspace{\fill}
		\subfloat[Covertype]{
			\includegraphics[width=0.3\textwidth]{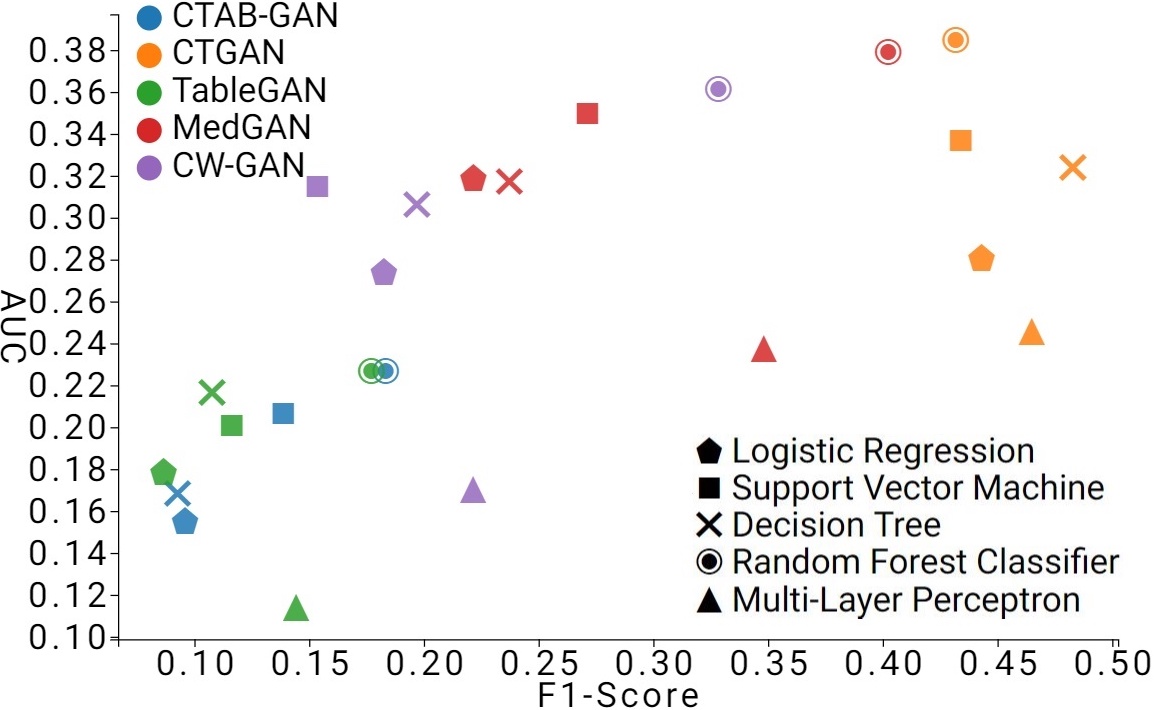}
			\label{fig:result_covtype}
 		}
	\hspace{\fill}
		\subfloat[Intrusion]{
			\includegraphics[width=0.3\textwidth]{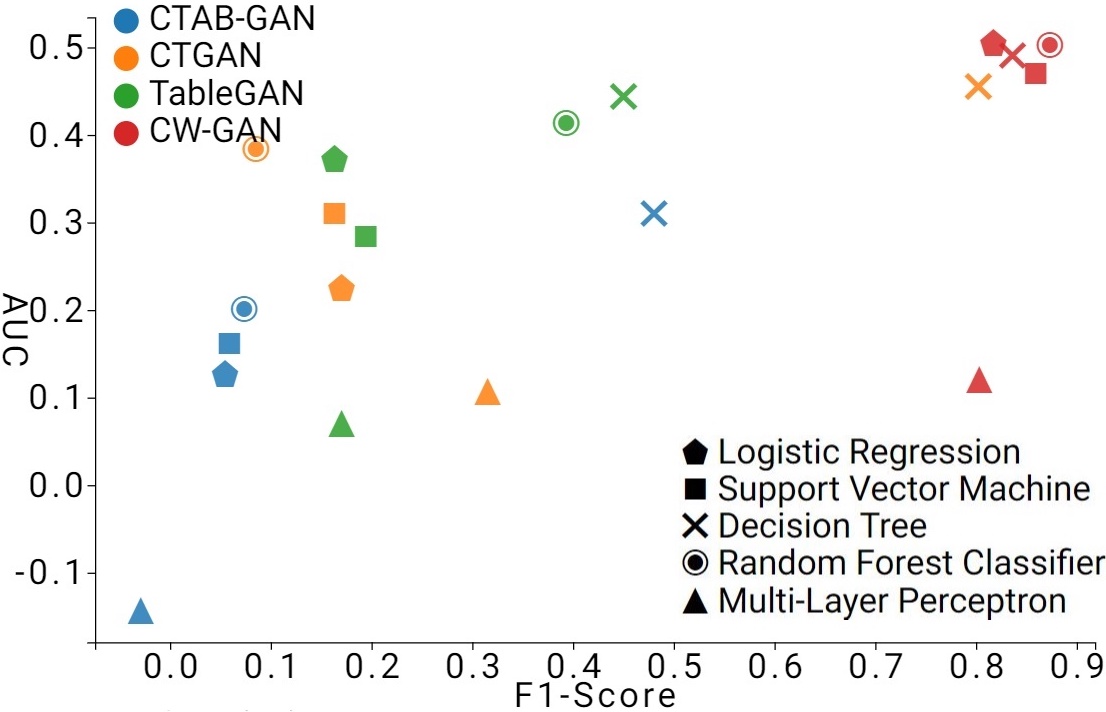}
			\label{fig:result_intrusion}
		}
	\quad
		\subfloat[Loan]{
			\includegraphics[width=0.3\textwidth]{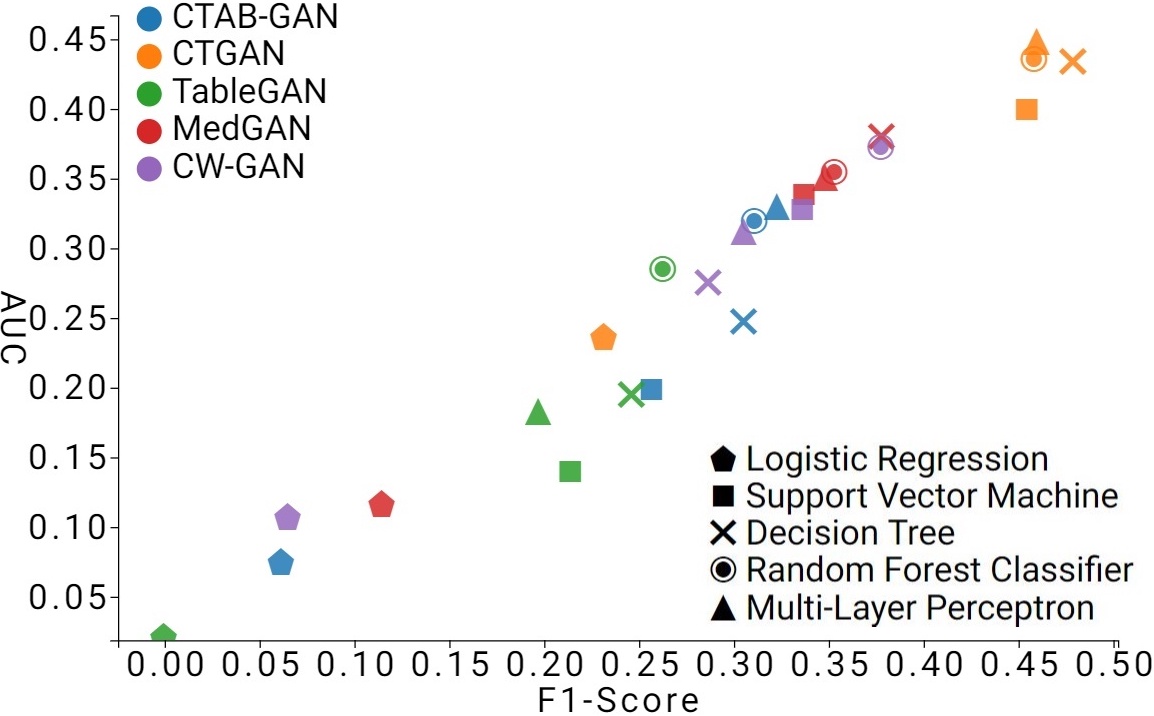}
			\label{fig:result_loan}
		}
		\caption{ML utilities difference (i.e., AUC and F1-scoree) for five algorithms based on five synthetically generated data}
		\label{fig:ml_whole}
	\end{center}
\end{figure*}

\textbf{ML Utility}. Tab.~\ref{table:ML_all} shows the averaged ML utility differences between real and synthetic data in terms of accuracy, F1 score, and AUC. A better synthetic data is expected to have low differences. It can be seen that \algo outperforms all other state-of-the-art methods in terms of Accuracy, F1-score and AUC. Accuracy is the most commonly used classification metric, but since we have imbalanced target \column, F1-score and AUC are more stable metrics for such cases. AUC ranges value from 0 to 1. \algo largely shortens the AUC difference from 0.169 (best in state-of-the-art) to 0.117.  

\begin{table}[t]
\centering
\caption{Difference of ML accuracy (\%), F1-score, and AUC between original and synthetic data: average over 5 different datasets and 3 replications.}
\begin{tabular}{|c|c|c|c|}
\hline
\textbf{Method} & \textbf{Accuracy} & \textbf{F1-score}  & \textbf{AUC} \\
\hline
\small{\algo}    & \textbf{9.83\%} & \textbf{0.127} &   \textbf{0.117}  \\
\small{CTGAN}    &21.51\% & 0.274 &   0.253 \\
\small{TableGAN} &11.40\%  & 0.130 &   0.169  \\
\small{MedGAN}    & 14.11\%&  0.282&   0.285 \\
\small{CW-GAN}    &  20.06\%& 0.354 &  0.299 \\
\hline
\end{tabular}
\label{table:ML_all}
\end{table}

To obtain a better understanding, Fig.~\ref{fig:ml_whole} plots the (F1-score, AUC) for all 5 ML models for the Covertype, Intrusion and Loan datasets. Due to the page limit restrictions, results for Adult and Credit datasets are not shown. Their results are similar as the ones of the Covertype dataset. Furthermore, Fig.~\ref{fig:result_intrusion} shows that for the Intrusion dataset \algo largely outperforms all others across all  ML models used for evaluation. For datasets such as Covertype, the results of \algo and \tablegan are similar and clearly better than the rest. This is because apart from \ctgan, the other models fail to deal with the imbalanced categorical \columns. Furthermore, as \ctgan uses a VGM model with 10 modes, it fails to converge to a suitable optimum for Covertype that mostly comprises single mode Gaussian distributions. 

For the Loan dataset, \tablegan is better than \algo and others, but the difference between the two is smaller than for the Intrusion dataset. We believe that the reason \algo outperforms the others by such a wide margin (17\% higher than second best for averaged accuracy across the 5 machine learning algorithms) for the Intrusion dataset is that it contains many highly imbalanced categorical \columns. In addition, Intrusion also includes 3 long tail continuous variables. Our results indicate that none of the state-of-the-art techniques can perform well under these conditions. The Loan dataset is significantly smaller than the other 4 datasets and has the least number of \columns. Moreover, all continuous \columns are either simple one mode Gaussian distributions or just uniform distributions. Therefore, we find that the encoding method in \algo which works well for complex cases, fails to converge to a better optimum for simple and small datasets. 

\begin{table}[t]
\centering
\caption{Statistical similarity: three measures averaged over 5 datasets and three repetitions.}
\begin{tabular}{|c|c|c|c|}
\hline
\textbf{Method} & \textbf{Avg JSD} & \textbf{Avg WD}  & \textbf{Diff. Corr.} \\
\hline
\small{\algo}   & \textbf{0.0697}&  \textbf{1050}  &  \textbf{2.10} \\
\small{CTGAN}    & 0.0704&  1769  & 2.73  \\
\small{TableGAN} & 0.0796& 2117   &  2.30 \\
\small{MedGAN}   & 0.2135& 46257  &   5.48   \\
\small{CW-GAN}   & 0.1318& 238155 &  5.82 \\
\hline
\end{tabular}
\label{table:SS_all}
\vspace{-0.5em}
\end{table}

\textbf{Statistical similarity}. Statistical similarity results are reported in Tab.~\ref{table:SS_all}. \algo stands out again across all comparisons. 
For categorical \columns (i.e. average JSD), \algo and \ctgan perform similarly (1\% difference),  
and better than the other methods by at least 12.4\%, i.e. against the next best \tablegan. This is due to the use of a conditional vector and the log-frequency sampling of the training data, which works well for both balanced and imbalanced distributions. For continuous \columns (i.e. average WD), we still benefit from the design of the conditional vector. The average WD column shows some extreme numbers such as 46257 and 238155 comparing to 1050 of \algo. The reason is that these algorithms generate extremely large values for long tail \columns. Besides divergence and distance, our synthetic data also maintains better correlation. We can see that \tablegan also performs well here. However, as the extended conditional vector enhances the training procedure, this helps to maintain even more so the correlation between \columns. This is because the extended conditional vector allows the generator to produce samples conditioned even on a given VGM mode for continuous \columns. This increases the capacity to learn the conditional distribution for continuous \columns and hence leads to an improvement in the overall feature interactions captured by the model.  

\textbf{Privacy preservability}. As only PATE-GAN can generate synthetic data within tight differential privacy guarantees, we only use distance-based algorithms to give an overview on privacy in our evaluation. On the one hand, if the distance between real and synthetic data is too large, it simply means that the quality of generated data is poor. On the other hand, if the distance between real and synthetic data is too small, it simply means that there is a risk to reveal sensitive information from the training data. Therefore, the evaluation of privacy is relative. The privacy results are shown in Tab.~\ref{table:PP_all}. It can be seen that the DCR and NNDR between real and synthetic data all indicate that generation from \tablegan has the shortest distance to real data (highest privacy risk).
The algorithm which allows for greater distances between real and synthetic data under equivalent ML utility and statistical similarity data should be considered. In that case, \algo not only outperforms \tablegan in ML utility and statistic similarity, but also in all privacy preservability metrics by 10.3\% and 4.6\% for DCR and NNDR, respectively. 
Another insight from this table is that for \medgan, DCR  within synthetic data is 41\% smaller than within real data. This suggests that it suffers from the mode collapse problem.


\begin{table}[t]
\centering
\caption{Privacy impact: between real and synthetic data (R\&S) and within real data (R) and synthetic data (S).}
\begin{tabular}{|c|c|c|c|c|c|c|}
\hline
\multirow{2}{*}{\textbf{Model}} & \multicolumn{3}{c|}{\textbf{DCR}} & \multicolumn{3}{c|}{\textbf{NNDR}} \\
\cline{2-7}
 & \textbf{R\&S} & \textbf{R}  & \textbf{S} & \textbf{R\&S} & \textbf{R}  & \textbf{S}\\

\hline
\algo     & 1.101 &  0.428& 0.877 & 0.714 & 0.414 &0.558\\
CTGAN    & 1.517 &  0.428& 1.026 & 0.763 & 0.414 &0.624 \\
TableGAN & 0.988 &  0.428& 0.920 & 0.681 & 0.414 &0.632\\
MedGAN   & 1.918 &  0.428& 0.254 & 0.871 & 0.414 &0.393 \\
CW-GAN   & 2.197 &  0.428& 1.124 & 0.847 & 0.414 &0.675\\
\hline
\end{tabular}
\label{table:PP_all}
\end{table}

\subsection{Ablation analysis}
To illustrate the efficiency of each strategy we implement an ablation study which cuts off the different components of \algo one by one:

\begin{itemize}[noitemsep]

\item[] \textbf{w/o classifier}. In this experiment, Classifier and the corresponding classification loss for Generator is taken away from \algo. 

\item[] \textbf{w/o information loss}. In this experiment, we remove information loss from \algo.

\item[] \textbf{w/o VGM and mode vector}. In this case, we substitute VGM for continuous \columns with min-max normalization and use simple one-hot encoding for categorical \columns. Here the conditional vector is the same as for \ctgan.
\end{itemize}


The results are compared with the baseline implementing all strategies. 
All experiments are repeated 3 times, and results are evaluated on the same 5 machine learning algorithms introduced in Sec.~\ref{sec:ml_efficacy}. 
The test datasets and evaluation flow are the same as shown in Sec.~\ref{ssec:setup} and Sec.~\ref{sec:metrics}. 
Tab.~\ref{table:ablation} shows the results.
Each part of \algo has different impacts on different datasets. For instance, \textbf{w/o classifier} has a negative impact for all datasets except Credit.
Since Credit has only 30 continuous \columns and one target \column, the semantic check can not be very effective.
\textbf{w/o information loss} has a positive impact for Loan, but results degenerate for all other datasets. It can even make the model unusable, e.g. for Intrusion. \textbf{w/o VGM and mode vector} performs bad for Covertype, but has little impact for Intrusion. Credit w/o VGM and mode vector performs  better than original \algo. This is because out of 30 continuous \columns, 28 are nearly single mode Gaussian distributed. The initialized high number of modes, i.e. 10, for each continuous \column (same setting as in \ctgan) degrades the estimation quality. 
In general, if we average the column values, all the ablation tests have a negative impact for the performance which justifies our design choices for \algo. 

\begin{table}[t]
\centering
\caption{F1-score difference to \algo. \algo column reports the absolute averaged F1-score as baseline.}
\label{table:ablation}
\resizebox{\columnwidth}{!}{
\begin{tabular}{ |C{1.4cm}|C{1.4cm}|C{1.2cm}|C{1.5cm}| C{1.5cm}|}
\hline
\textbf{Dataset} & \textbf{w/o Classifier} & \textbf{w/o Info. Loss}  & \textbf{w/o VGM and Mode vector}& \textbf{\algo} \\
\hline
Adult &  -0.01&	-0.037 & -0.05  & 0.704  \\
\hline
Covertype & -0.018 & -0.184& -0.118  & 0.532\\
\hline
Credit & +0.011& -0.177& +0.06&  0.71 \\
\hline
Intrusion &-0.031&-0.437&+0.003& 0.842 \\
\hline
Loan &-0.044&+0.028&+0.013  & 0.803 \\
\hline
\end{tabular}
}
\end{table}


\begin{figure*}[t]
	\begin{center}
		\subfloat[Mortgage in Loan]{
			\includegraphics[width=0.33\textwidth]{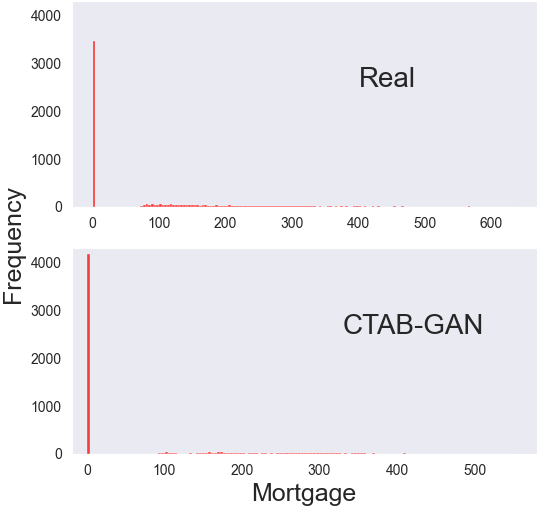}
			\label{fig:ctabgan_bimodal}
		}
		\subfloat[Amount in Credit]{
			\includegraphics[width=0.31\textwidth]{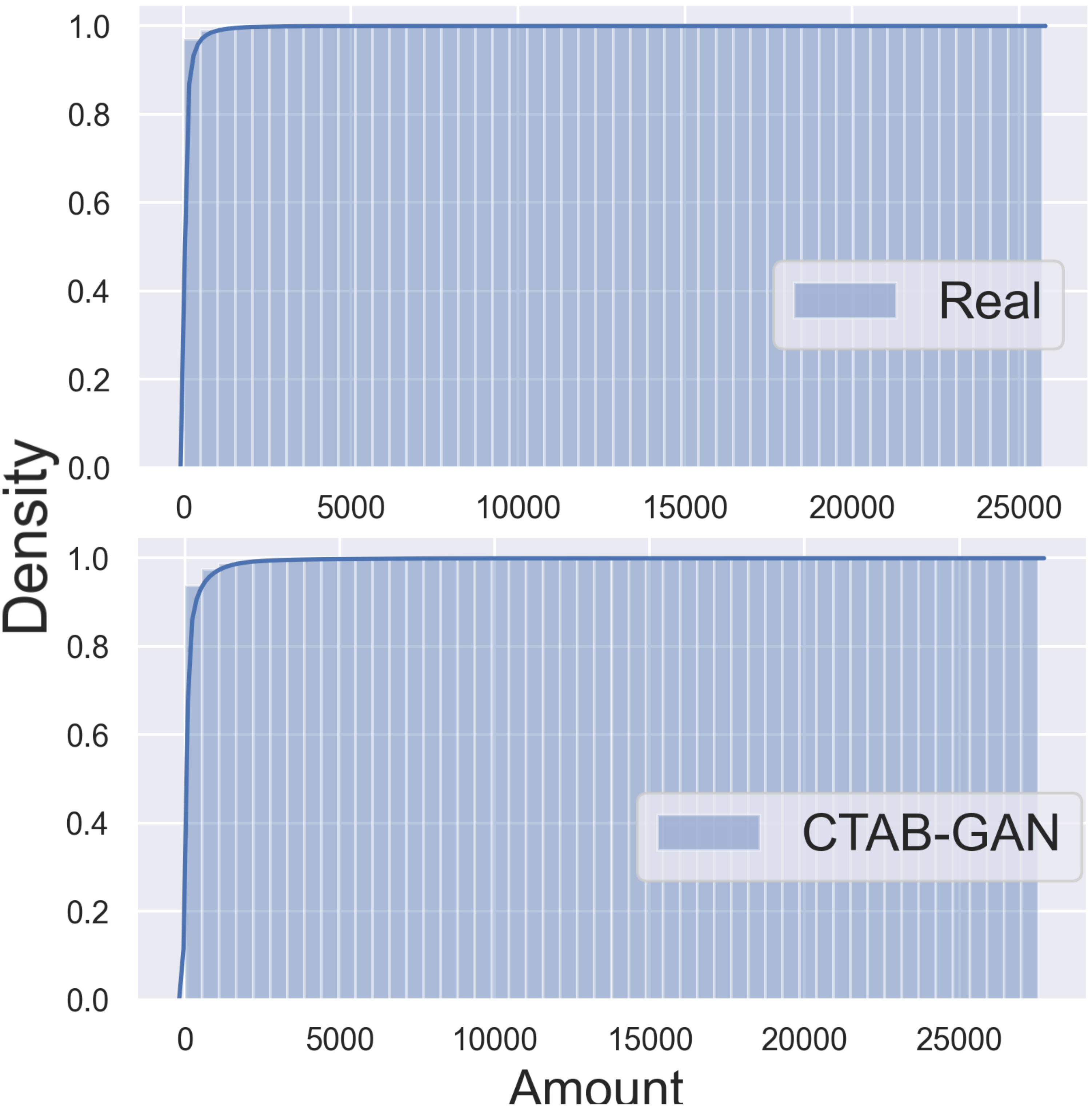}
			\label{fig:longtail_result}
		}
		\subfloat[Hours-per-week in Adult]{
			\includegraphics[width=0.335\textwidth]{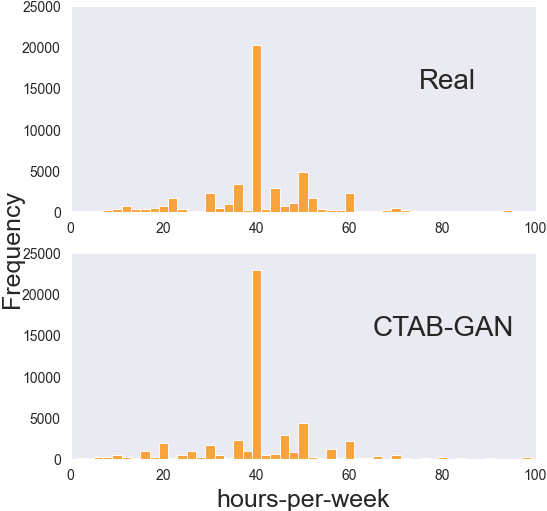}
			\label{fig:gmm_result}
		}
		\caption{ Challenges of modeling industrial dataset using existing GAN-based table generator: (a) \bimodal type, (b) long tail distribution, and (c) \skewed data} 
		\label{fig:motivationcases_response}
 	\end{center}
 	\vspace{-0.5em}
\end{figure*}

\subsection{Further discussion}
\label{sec:further}
After reviewing all the metrics, let us recall the three motivation cases from Sec.~\ref{sec:motivation}. 

{\bf \Mixed data type \variables}. Fig.~\ref{fig:ctabgan_bimodal} compares the real and \algo generated data for \column \textit{Mortgage} in the Loan dataset. \algo encodes this \column as \bimodal type. We can see that \algo generates clear 0 values. One drawback which can be observed is that \algo amplifies the dominance of the 0 in this \column. Even with our log-frequency-based sampling of Gaussian mixture modes and categorical classes, \algo generates more 0 values than in the original distribution. That means there is still room for improvement for extremely imbalanced cases. 

{\bf Long tail distributions.} Fig.~\ref{fig:longtail_result} compares the cumulative frequency graph for the \textit{Amount} \column in Credit. This \column is a typical long tail distribution. 
One can see that \algo perfectly recovers the real distribution. Due to log-transform data pre-processsing, \algo learns this structure significantly better than the state-of-the-art methods shown in Fig.~\ref{fig:amount_result_motivation}.  

{\bf \Skewed multi-mode continuous \variables}.
Fig.~\ref{fig:gmm_result} compares the frequency distribution for the continuous \column \textit{Hours-per-week} from Adult. Except the dominant peak at 40, there are many side peaks. Fig.~\ref{fig:gmm_result_motivation}, shows that \tablegan, \cwgan and \medgan struggle since they can learn only a simple Gaussian distribution due to the lack of any special treatment for continuous \columns. \ctgan, which also use VGM, can detect other modes. Still, \ctgan is not as good as \algo. The reason is that \ctgan lacks the mode of continuous \columns in the conditional vector. By incorporating the  mode of continuous \columns into conditional vector, we can apply the training-by-sample and logarithm frequency also to modes. This gives the mode with less weight more chance to appear in the training and avoids the mode collapse. 

\section{Conclusion}

Motivated by the importance of data sharing and fulfillment of governmental regulations, we propose \algo -- a conditional GAN based tabular data generator. \algo advances beyond the prior state-of-the-art methods by modeling \bimodal \variables and provides strong generation capability for imbalanced categorical \variables, 
and continuous \variables with complex distributions.
To such ends, the core features of \algo include (i) introduction of the classifier into conditional GAN, (ii) effective data encoding for \bimodal \variable, and (iii) a novel construction of conditional vectors. We exhaustively evaluate \algo against four tabular data generators on a wide range of metrics, namely resulting ML utilities, statistical similarity and privacy preservation. The results show that the synthetic data of \algo results into  high utilities, high similarity and reasonable privacy guarantee, compared to existing state-of-the-art techniques. The improvement on complex datasets is up to 17\% in accuracy comparing to all state-of-the-art algorithms. 
The remarkable results of \algo demonstrate its potential for a wide range of applications that greatly benefit from data sharing, such as banking, insurance, manufacturing, and telecommunications. 

\ifnotdoubleblind
\section*{Acknowledgements}
This work has been funded by Ageon data science division.
\fi

\bibliographystyle{abbrv}
\bibliography{sigproc}  
\ifnotdoubleblind
\pagebreak
\appendix
\renewcommand\thefigure{\thesection.\arabic{figure}}   
\section{User Manual}
\setcounter{figure}{0} 
\subsection{Introduction}

The software demo developed by us comprises of a synthetic tabular data generation pipeline. It was implemented using python 3.7.* along with the flask library to work as a web application on a local server. The application functionality and usage can be found listed under the \nameref{Fmarker} \& \nameref{Umarker} sections respectively. In addition, a video of the demo can be seen \href{https://drive.google.com/file/d/1VK6479YPnjg0zVbfdgJb2G4_7lz2CWp6/view}{\underline{here}}. 

\subsection{Functionality}
\label{Fmarker}

Our demo comprises of the following salient features:

\begin{enumerate}
    
    \item \textbf{Synthetic Data Generator: }Our software is a cross-platform application that sits on top of a python interpreter. Moreover, it is relatively lightweight and can be set-up easily using pip. 
    Our application is also robust against missing values and supports date-type formats. We believe these factors increases its usability in real-world scenarios.  
    
    \item \textbf{Synthetic Data Evaluator: }In addition to our generator, we also provide a detailed evaluation of the synthetic data. The report provides end users with visual plots comparing the real and synthetic distributions of individual columns as shown in sub-figures \ref{fig:workclass} \& \ref{fig:age}  of Fig.~\ref{fig:visual_plots}.  In addition, the synthetic data's utility for ML applications along with its privacy preservability metrics are reported as can be seen in sub-figures \ref{fig:utility} \& \ref{fig:privacy} of Fig.~\ref{fig:efficacy}. Note that the table-evaluator\footnote{\url{ttps://github.com/Baukebrenninkmeijer/Table-Evaluator}} library aided us in generating this evaluating report.

\end{enumerate}

\subsection{Usage}
\label{Umarker}

The following step-by-step instructions are provided to allow end-users to use our product in a hassle-free manner. 

\begin{enumerate}[start=1,label={\bfseries Step \arabic*:},leftmargin=1.425cm]
    \item Open the terminal and navigate to the root directory of the software package to run the following command \texttt{python3 / python server.py}. 
    \item  \begin{minipage}[t]{\linewidth}
          \raggedright
          \adjustbox{valign=t}{%
            \includegraphics[width=.8\linewidth]{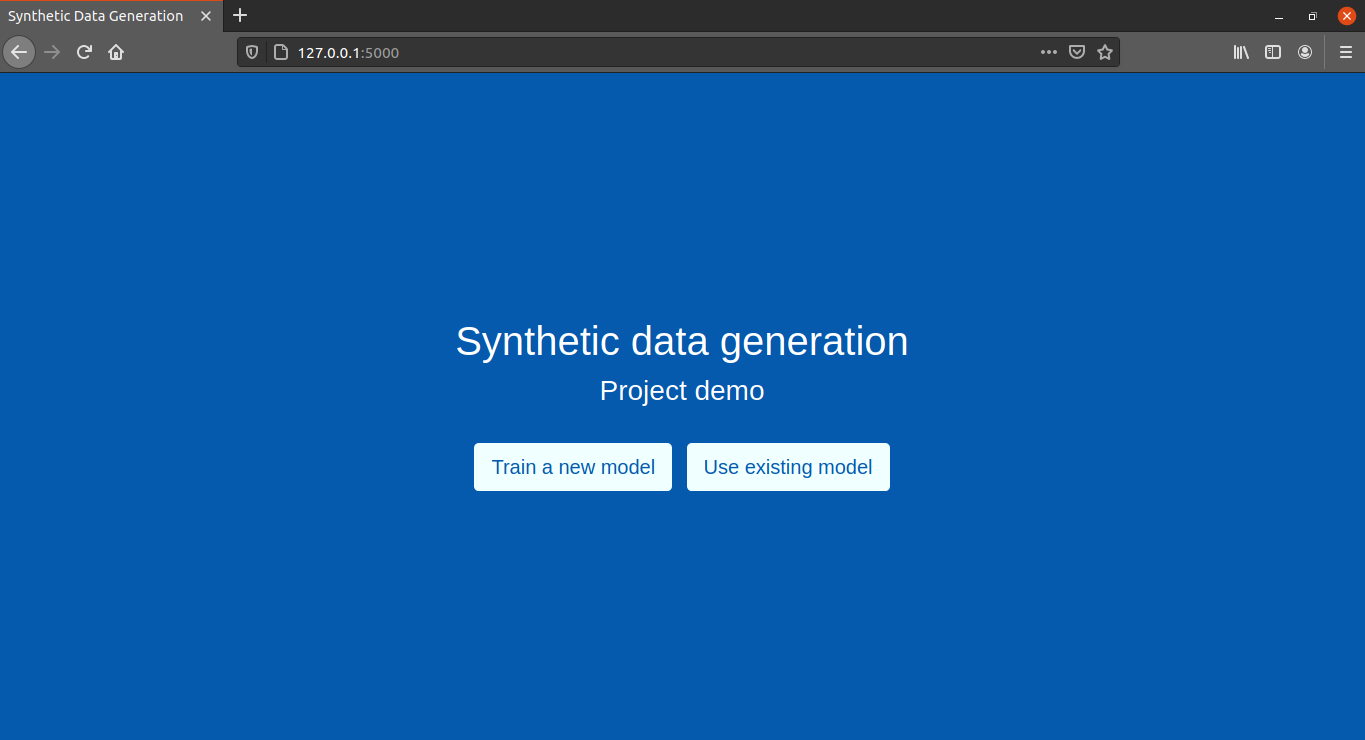}%
          }
          \medskip
          
          Open the browser, the application should now be available at the following address: \texttt{http://127.0.0.1:5000/}.

    \end{minipage}
    
    \item If this is the first time running the web application, it is advised to click on the ``Train a new model'' button to begin training the model with a dataset. Otherwise, click on the ``Use existing mode'' button to use an existing trained model.
    If you clicked on the ``Use existing mode'' button, please go to \textbf{step 8}. If not, please continue with the next step.
    
    \item \begin{minipage}[t]{\linewidth}
          \raggedright
          \adjustbox{valign=t}{%
            \includegraphics[width=.8\linewidth]{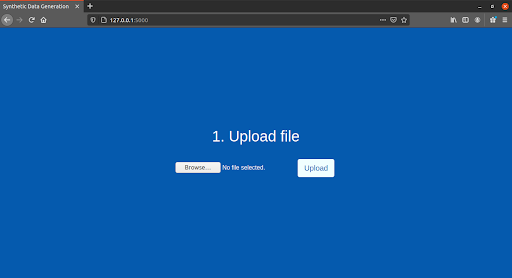}%
          }
          \medskip
    
     Click on the ``Browse'' button to select the dataset for which the model needs to train. Afterwards click on the ``Uploap'' button. 
     
    \end{minipage}

    \item \begin{minipage}[t]{\linewidth}
          \raggedright
          \adjustbox{valign=t}{%
            \includegraphics[width=.8\linewidth]{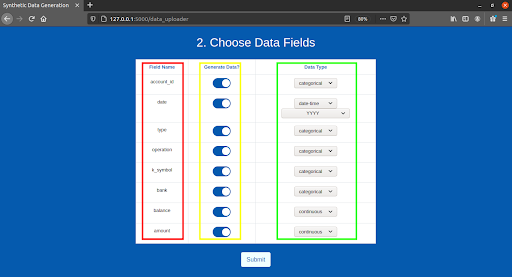}%
          }
          \medskip
    
     The software will auto-detect the column types, and will give the option to adjust a column's data type and inclusion in the training. Note that the red highlighted column shows the current columns in the uploaded \texttt{csv} file. The yellow highlighted column gives the option to include or exclude a particular column in the training process by clicking on the switch button. The highlighted green column is the auto detected data type. It also has the option to be adjusted as needed. Simply click on it and select the desired data type from the drop down menu. Click on the ``Submit'' button after choosing the right settings.
    \end{minipage}
    
    \item \begin{minipage}[t]{\linewidth}
          \raggedright
          \adjustbox{valign=t}{%
            \includegraphics[width=.8\linewidth]{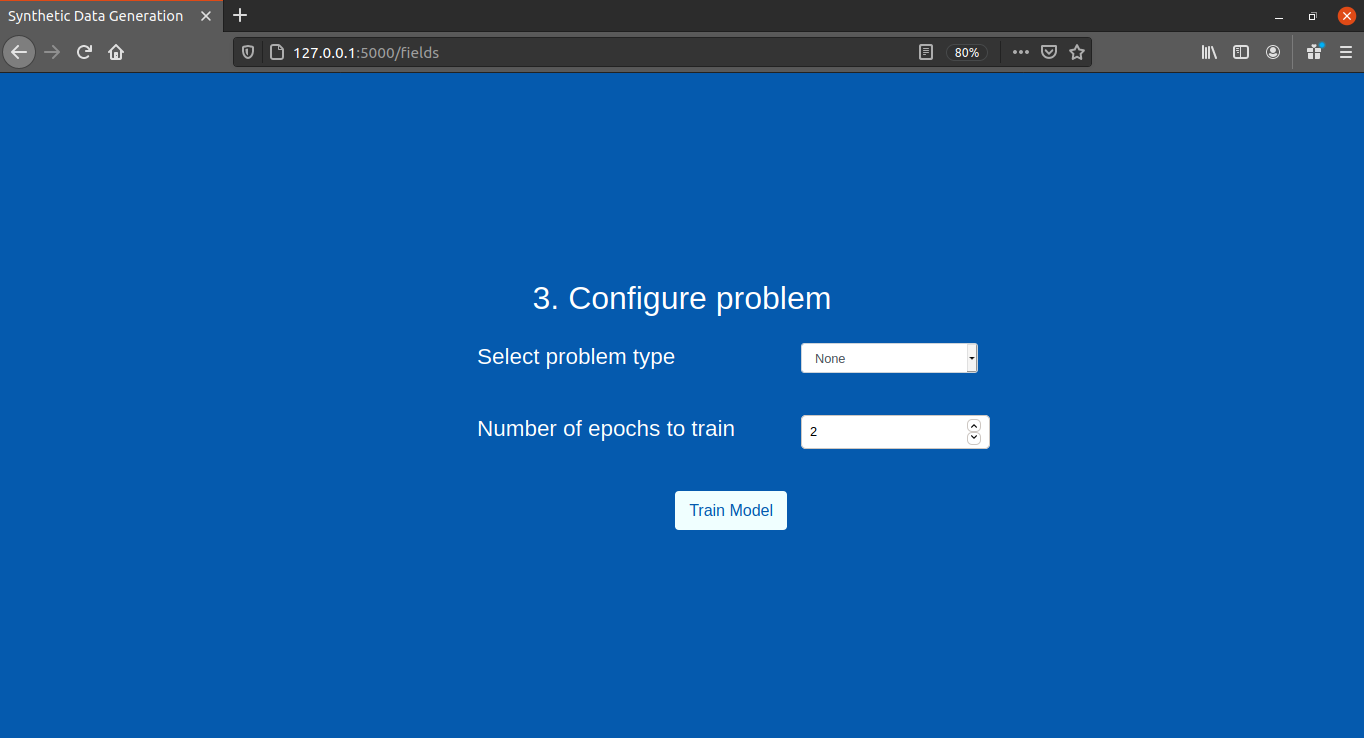}%
          }
          \medskip
    
     In the following page, specify the problem type for the given dataset. The software currently provides the following problem types: None, Binary Classification and Multi-class Classification. If unsure, leave it as None. Then enter the number of epochs needed to train the model. Click on ``Train Model'' to start the training. 

    \end{minipage}
    
    \item \begin{minipage}[t]{\linewidth}
          \raggedright
          \adjustbox{valign=t}{%
            \includegraphics[width=.8\linewidth]{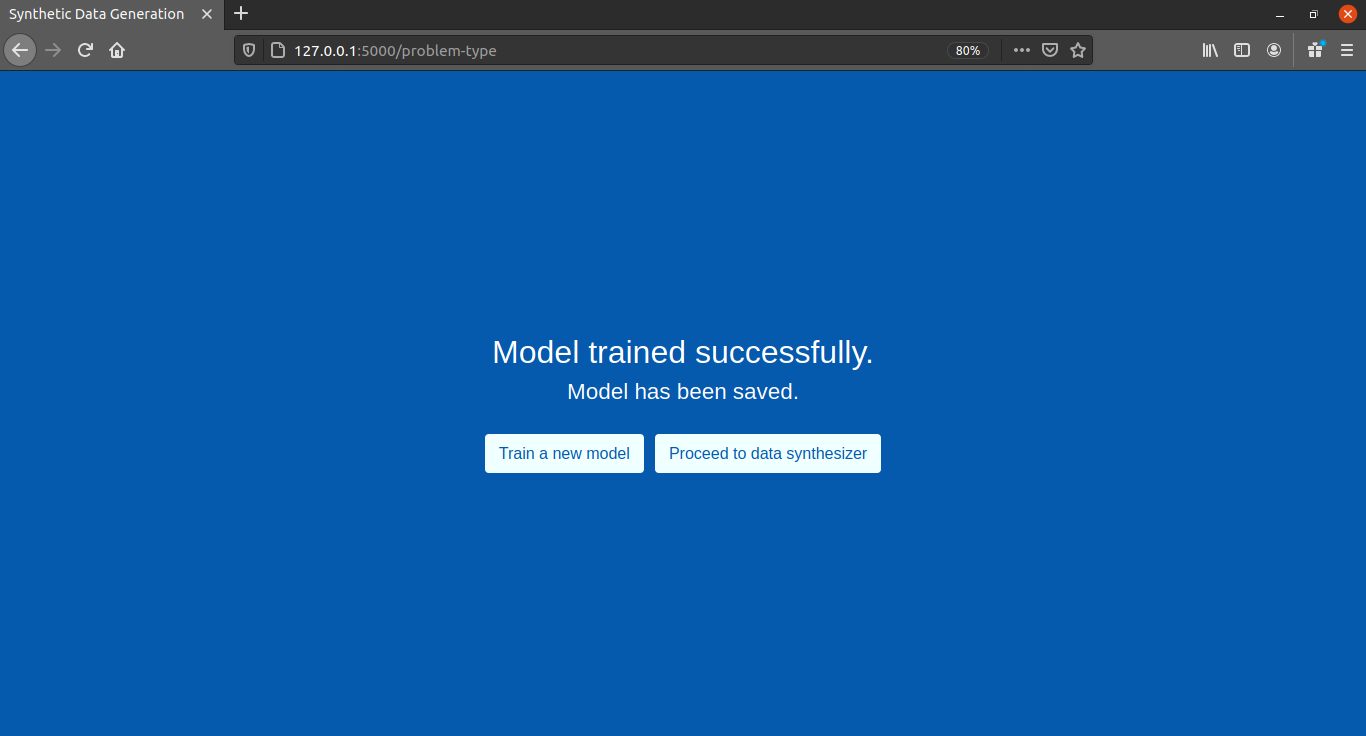}%
          }
          \medskip
    
     Once the model has finished training, the option to train a new model or proceed to the synthesizer is presented. To generate synthetic data, click on ``Proceed to data synthesizer''.

    \end{minipage}
    
    \item \begin{minipage}[t]{\linewidth}
          \raggedright
          \adjustbox{valign=t}{%
            \includegraphics[width=.8\linewidth]{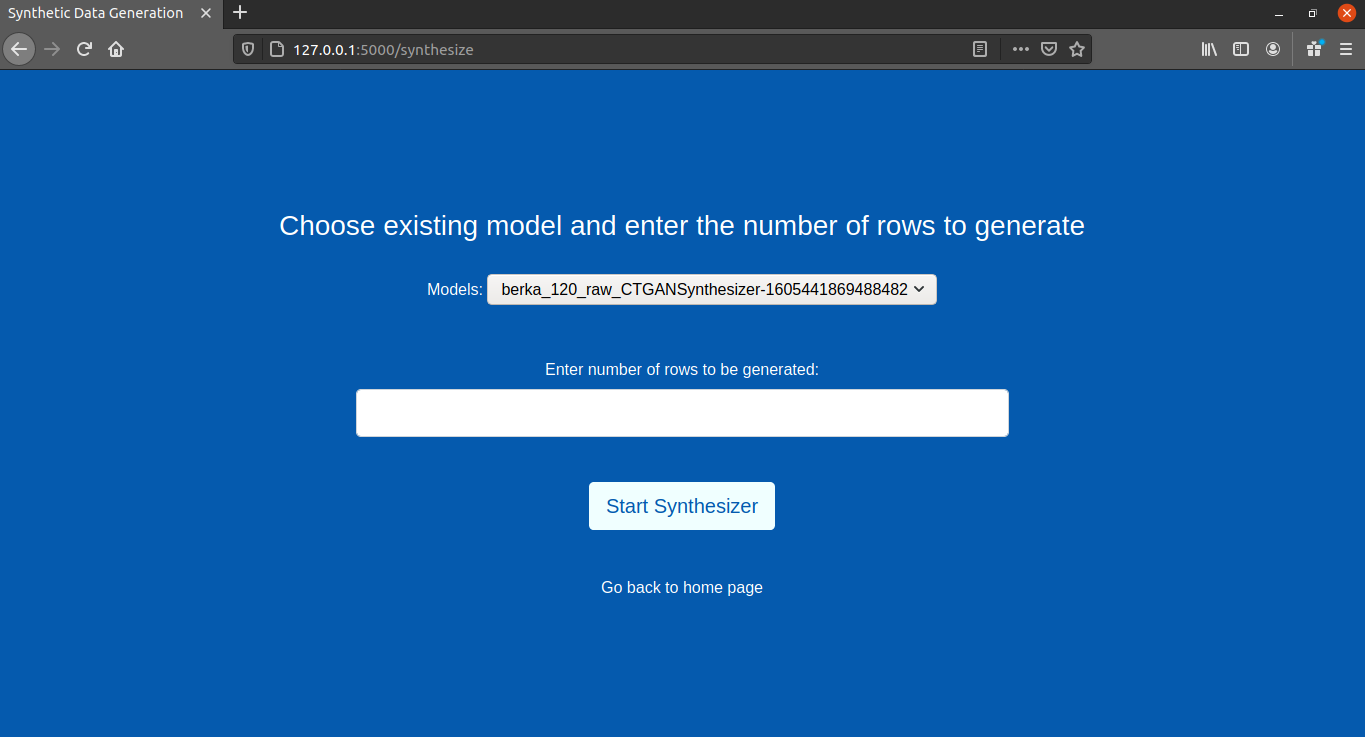}%
          }
          \medskip
    
     The trained models can be found in the dropdown menu of the Models field, as seen in the figure above. Click on it, and select the trained model. After this, type the amount of rows to be generated in the second field, and click on ``Start Synthesizer'' to start the process. 

    \end{minipage}
    
    \item \begin{minipage}[t]{\linewidth}
          \raggedright
          \adjustbox{valign=t}{%
            \includegraphics[width=.8\linewidth]{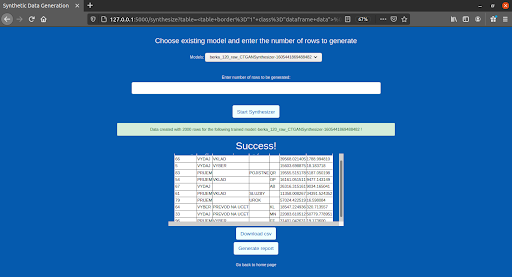}%
          }
          \medskip
    
     Once the data is generated, the following page shows a snippet of the synthetic data. The generated data can be saved locally by clicking on ``Download csv''. This page also gives you the option to generate a report for the given data. In order to generate the report in PDF format, simply click on ``Generate report'' and continue with step 10. 

    \end{minipage}
    
    \item \begin{minipage}[t]{\linewidth}
          \raggedright
          \adjustbox{valign=t}{%
            \includegraphics[width=.8\linewidth]{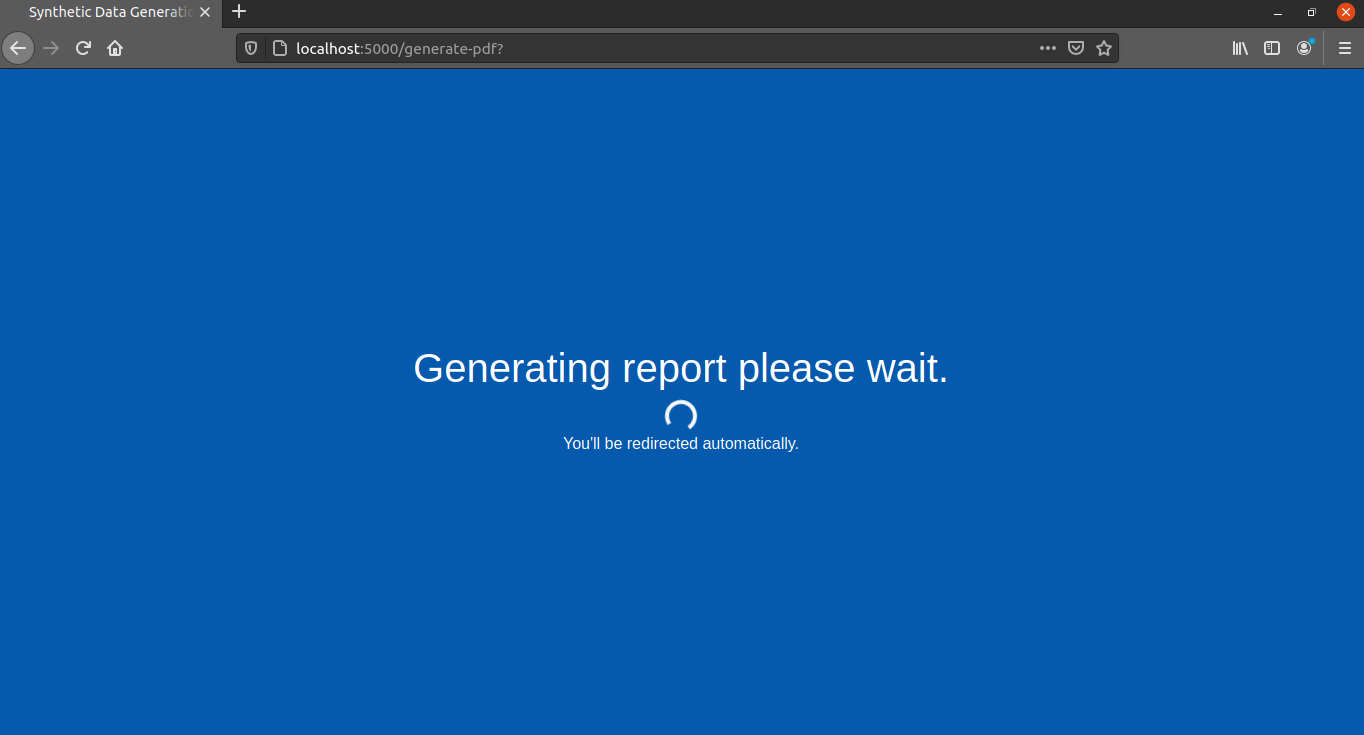}%
          }
          \medskip
    
    The following page is presented while the report is being generated. It will automatically redirect to the PDF once it is completed.
    \end{minipage}
    
    \item \begin{minipage}[t]{\linewidth}
          \raggedright
          \adjustbox{valign=t}{%
            \includegraphics[width=.8\linewidth]{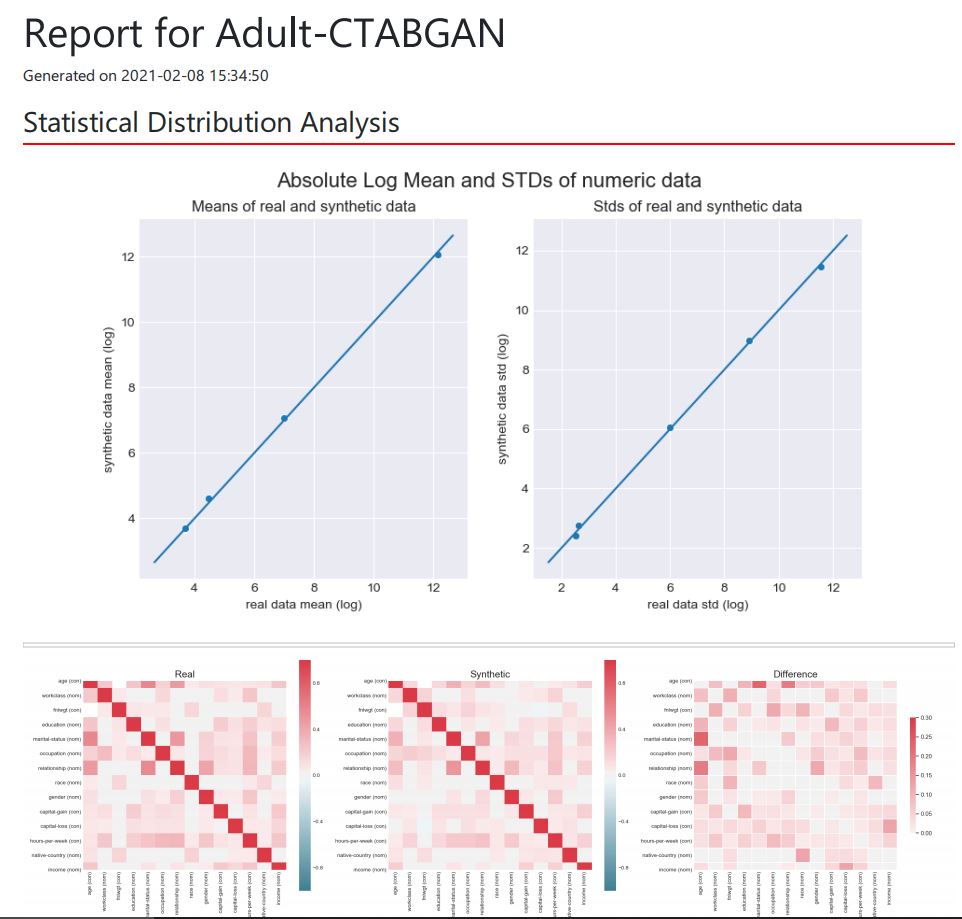}%
          }
          \medskip
    
    Finally, once the PDF has been generated, it can be saved locally by clicking on ``Save as'' or ``Print'' in the browser.  
    \end{minipage}

\end{enumerate}

\begin{figure}[H]
	\begin{center}
		\subfloat[Cumulative distribution comparison of Age in Adult]{
			\includegraphics[width=0.47\columnwidth,height=4.5cm]{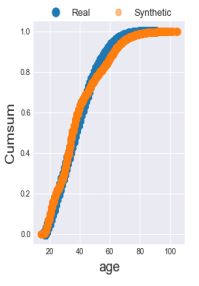}
		    \label{fig:workclass}
		}
		\hfil
		\subfloat[Frequency comparison of categories within Workclass in Adult]{
			\includegraphics[width=0.47\columnwidth,height=4.25cm]{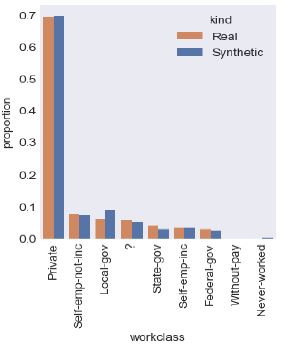}
			\label{fig:age}
		}
		\caption{Visual plots comparing the generated vs real data distribution.}
		\label{fig:visual_plots}
	\end{center}
\end{figure}

\begin{figure}[H]
	\begin{center}
		\subfloat[ML Utility]{
			\includegraphics[width=0.57\columnwidth]{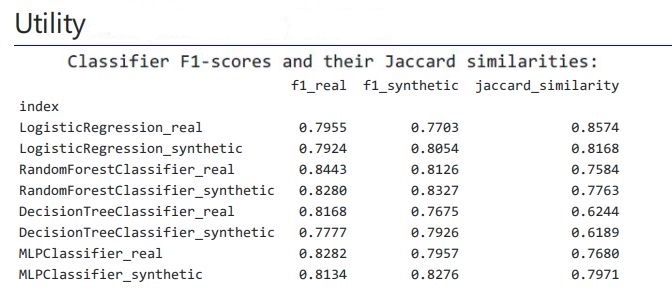}
		    \label{fig:utility}
		}
		\hfil
		\subfloat[Privacy Preservability]{
			\includegraphics[width=0.57\columnwidth]{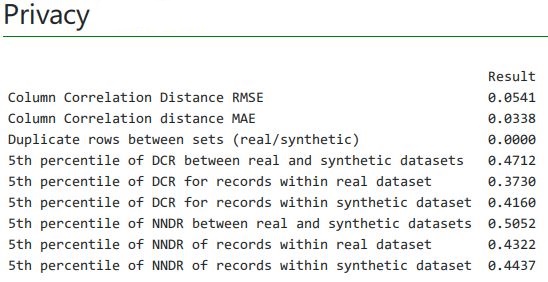}
			\label{fig:privacy}
		}
		\caption{ML utility and privacy preservability of the generated data.}
		\label{fig:efficacy}
	\end{center}
\end{figure}

\fi

\end{document}